\PassOptionsToPackage{table,dvipsnames,rgb}{xcolor}
\documentclass{maskwam}

\usepackage[utf8]{inputenc}
\usepackage{amsmath,amssymb}
\usepackage{makecell}
\usepackage{pifont}
\newcommand{\cmark}{\ding{51}}
\newcommand{\xmark}{\ding{55}}

\title{Follow-Bench 2.0: An End-to-End 3D Benchmark for Socially-Aware Robot Person Following}
\author[1,2,*]{Hanjing Ye}
\author[1,*]{Tianle Zeng}
\author[1,*]{Jianwei Peng}
\author[1]{Yanci Wen}
\author[2]{Yuchen Zhou}
\author[2,3,\#]{Yonggen Ling}
\author[1,\#]{Hong Zhang}
\affiliation[1]{Southern University of Science and Technology}
\affiliation[2]{Tencent Robotics X}
\affiliation[3]{Futian Laboratory}
\contribution[*]{Equal contribution}
\metadata[Project page]{\url{https://medlartea.github.io/follow-bench-v2/}}
\metadata[Code]{\url{https://github.com/MedlarTea/follow-bench-2.0/}}
\correspondence{\email{rolandling@tencent.com}, \email{hzhang@sustech.edu.cn}}

\abstract{
  Socially-aware robot person following (RPF) requires a mobile robot to follow a designated person in dynamic human environments while maintaining target identity, avoiding obstacles and pedestrians, and preserving socially comfortable formations. Existing benchmarks only partially evaluate this coupled problem: embodied visual tracking benchmarks emphasize target visibility with limited social interaction, person re-identification benchmarks study perception without closed-loop control, and motion-planning benchmarks often assume ground-truth target and pedestrian states. We introduce \textbf{Follow-Bench~2.0}, an end-to-end 3D benchmark for evaluating socially-aware RPF under coupled perception--planning challenges. Built in Unreal Engine, Follow-Bench~2.0 provides difficulty-leveled scenarios with diverse pedestrian flows, lighting and weather conditions, cluttered layouts, bottlenecks, queueing behaviors, distractors, and temporary target occlusions. The benchmark supports modular perception--planning pipelines, end-to-end visual policies, and foundation-model-based active trackers under a unified closed-loop protocol. It further evaluates both back- and side-following configurations and reports safety--comfort metrics covering task success, target visibility, target recovery, social-space intrusion, following formation, and motion smoothness. Experiments show that RPF-oriented modular pipelines substantially outperform current end-to-end visual policies on close and socially comfortable following, but also reveal that socially-aware RPF remains far from solved in interaction-heavy environments and especially in side-following. Follow-Bench~2.0 therefore provides a diagnostic platform for studying how perception errors, occlusions, crowd interactions, and following configurations jointly affect robot person following.
}

\begin{document}
\maketitle

\suppressfloats[t]
\begin{figure}[t]
  \centering
  \includegraphics[width=0.94\linewidth]{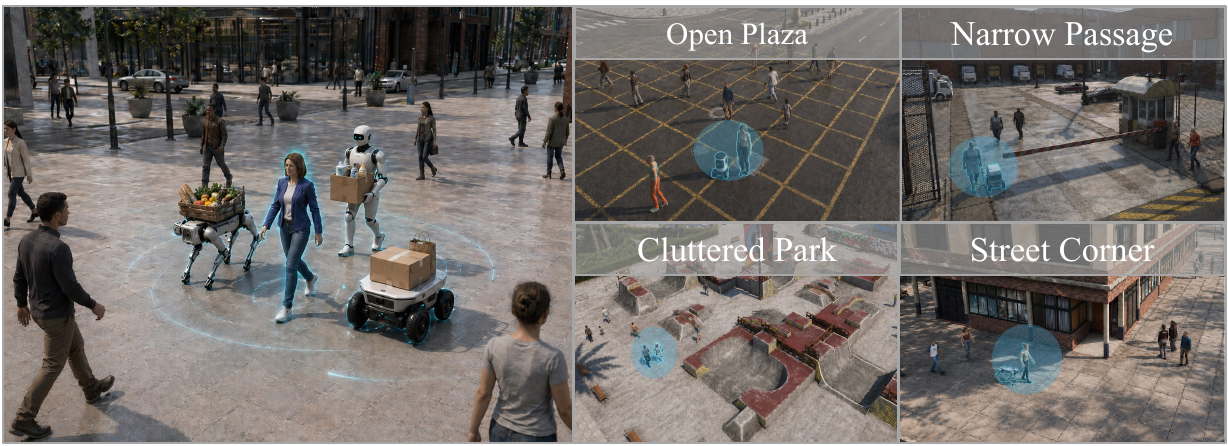}
  \captionsetup{width=0.94\linewidth}
  \caption{\textbf{Follow-Bench~2.0} evaluates socially-aware robot person following in photorealistic 3D crowds, with back- and side-following tasks across four diagnostic scenarios.}
  \label{fig:teaser}
\end{figure}

\section{Introduction}
\label{sec:intro}

Socially-aware robot person following (RPF)~\cite{islam2019person,honig2018toward,zhong2021advat,zhong2024empowering,wang2025trackvla} is a core capability for human-centered mobile robots in logistics, security patrol, shopping guidance, personal assistance, and companionship. Unlike generic visual tracking~\cite{muller2018trackingnet,fan2021lasot} or obstacle avoidance~\cite{han2025neupan,xie2023drl}, RPF requires a robot to continuously follow a designated person in dynamic human environments while preserving target identity, avoiding pedestrians and obstacles, and maintaining socially comfortable formations. This makes RPF a tightly coupled perception--planning problem: the robot must track the target under viewpoint changes, illumination variations, distractors, and occlusions, while simultaneously deciding where and how to move around surrounding people in a socially acceptable manner.

Despite growing interest in this problem, existing benchmarks only evaluate partial aspects of socially-aware RPF. Embodied visual tracking (EVT) benchmarks~\cite{wang2025trackvla,qiu2017unrealcv,sunopen} study target following in visually realistic environments, but mainly focus on keeping a target visible from behind and provide limited evaluation of social comfort and crowd interaction. Person tracking and re-identification (ReID) benchmarks~\cite{ye2025tpt,zuo2025reid5o} evaluate perception in isolation, without closing the loop between target localization and robot motion. Conversely, socially-aware motion-planning benchmarks such as Follow-Bench~1.0~\cite{ye2025rpf} evaluate safety and comfort in crowded environments, but assume ground-truth target and pedestrian states in a 2D simulator. As a result, current evaluations cannot systematically reveal how perception errors, occlusions, crowd interactions, environmental layouts, and following configurations jointly affect closed-loop RPF performance.

This gap is particularly important for learning-based RPF policies. Recent vision-language-action (VLA) policies~\cite{wang2025trackvla,zhang2025uni,liu2026comatrack,liu2025trackvla++,zhang2025embodied} aim to map raw visual observations directly to robot actions, but they are often trained through behavior cloning on expert trajectories. Since large-scale real-world RPF data is expensive to collect, risky to execute, and difficult to diversify across crowd behaviors and social contexts, simulation has become a practical source of training and evaluation episodes. However, reducing the sim-to-real gap requires more than photorealistic rendering. The simulated task distribution should expose policies to dense human--human and human--robot interactions, pedestrian- and obstacle-induced occlusions, diverse layouts, and social conventions that characterize everyday service scenarios. These coupled perception--planning challenges remain underrepresented in existing RPF benchmarks.

To address this gap, we introduce \textbf{Follow-Bench~2.0}, an end-to-end 3D benchmark for socially-aware RPF in photorealistic, human-populated, and socially interactive environments. Built upon Unreal Engine~\cite{qiu2017unrealcv,zhong2025unrealzoo,ren2025simworld,dosovitskiy2017carla,zeng2026leadvla}, Follow-Bench~2.0 does not claim simulator-level novelty. Instead, it focuses on RPF-challenge-driven benchmark design, providing scenarios, baselines, metrics, and interaction tools tailored to closed-loop person following in dynamic crowds. Our main contributions are summarized as follows:

1) \textbf{RPF-challenge-driven 3D benchmark design.}
Follow-Bench~2.0 provides difficulty-stratified 3D environments covering open plazas, narrow passages, cluttered parks, and street corners. These environments include diverse pedestrian flows, appearances, scene scales, weather and lighting conditions, and human--human/human--robot interactions. The scenarios are designed to stress key RPF challenges, including target ReID under occlusion and low lighting, pedestrian and obstacle avoidance, back- and side-following, and recovery after temporary target loss. The benchmark supports ground-truth perception, modular perception--planning pipelines, and end-to-end visual policies, with metrics covering target tracking, occlusion handling, collision avoidance, social comfort, following formation, and motion smoothness.

2) \textbf{Unified evaluation of modular and end-to-end baselines.}
We reproduce strong modular planners and representative end-to-end visual policies under a unified closed-loop protocol, enabling systematic comparison between classical perception--planning pipelines, end-to-end policies, and foundation-model-based active trackers. This comparison is important because many end-to-end following policies rely on expert trajectories generated by classical planning components, such as graph search, trajectory tracking, or controllers with privileged state information. The reproduced modular baselines can thus serve as evaluation references, expert demonstrators, distillation targets, and trajectory initializers for future imitation and reinforcement learning~\cite{qureshi2021motionplanningnetworks,carius2020mpcnet}.

3) \textbf{Extensible platform with visualization and user-interaction tools.}
Follow-Bench~2.0 provides visualization tools for tracked pedestrians, occupancy maps, planned trajectories, and other debugging signals. It also supports user-interactive control, allowing users to manually control the target person to test policies under customized behaviors, or control the robot to collect expert demonstrations. The platform includes multiple robot embodiments, including quadruped, humanoid, and wheeled robots, as well as common sensors such as RGB cameras, depth cameras, and LiDAR. We will open-source the platform to facilitate future research on socially-aware RPF, end-to-end visual navigation, and human-centered robot companion systems.

\paragraph{Key findings.}
Through closed-loop evaluation on Follow-Bench~2.0, we identify two findings that are difficult to observe in visibility-only or ground-truth-perception benchmarks. First, RPF-oriented modular pipelines substantially outperform current EVT-style baselines, including TrackVLA~\cite{wang2025trackvla} and OA-VAT~\cite{sun2026instance}, on close and socially comfortable following. This suggests that keeping the target visible from a distance is not sufficient for socially-aware RPF. Second, socially-aware RPF remains far from solved in interaction-heavy environments and in side-following configurations. In scenarios with narrow passages and rich human--human interactions, modular policies still suffer severe performance drops, while existing EVT-style baselines do not naturally support side-following. These results position socially-aware RPF as a coupled perception--planning benchmark with substantial headroom for both modular systems and future VLA/EVT policies.
\section{Related Work}
\label{sec:related}

\begin{table}[t]
  \centering
  \caption{Comparison of Follow-Bench~2.0 with embodied tracking and RPF-related benchmarks.}
  \label{tab:bench-comparison}
  \renewcommand{\arraystretch}{1.5}
  \setlength{\tabcolsep}{10pt}
  \resizebox{\linewidth}{!}{%
  \begin{tabular}{l c c c l c c l l l}
  \toprule
  \multirow{2}{*}{\textbf{Benchmark}}
    & \multicolumn{3}{c}{\textbf{RPF Capability Challenge}}
    & \multirow{2}{*}{\textbf{Follow Conf.}}
    & \multicolumn{2}{c}{\textbf{Eval. Baselines}}
    & \multirow{2}{*}{\textbf{Eval. Type}}
    & \multirow{2}{*}{\textbf{Ped. Inter.}}
    & \multirow{2}{*}{\textbf{Engine}} \\
  \cline{2-4}\cline{6-7}
  \noalign{\vskip 2.5pt}
    & \makecell[c]{Target\\ReID}
    & \makecell[c]{Obstacle /\\Occlusion Avoid.}
    & \makecell[c]{Socially-Aware\\Following}
    & & Modular & E2E & & & \\
  \midrule
  \multicolumn{10}{l}{\emph{Embodied visual-tracking benchmarks: mainly back-following with limited social interaction.}}\\
  \midrule
  EVT-Bench~\cite{wang2025trackvla}
    & Mid
    & Mid
    & Low
    & Back
    & \xmark & \cmark
    & task-level
    & \makecell[l]{Optimal Reciprocal\\Collision Avoidance\\(ORCA)*~\cite{van2011reciprocal}}
    & Habitat\,3.0 \\
  Gym-UnrealCV~\cite{qiu2017unrealcv}
    & Low
    & Low
    & Low
    & Back
    & \xmark & \cmark
    & task-level
    & NavMesh~\cite{dosovitskiy2017carla}
    & Unreal Engine \\
  DAT (aerial)~\cite{sunopen}
    & Low
    & Low
    & Low
    & Back
    & \xmark & \cmark
    & task-level
    & NavMesh
    & Unreal Engine \\
  \midrule
  \multicolumn{10}{l}{\emph{RPF-related benchmarks: RPF-relevant capabilities are evaluated only partially.}}\\
  \midrule 
  TPT-Bench~\cite{ye2025tpt}
    & High
    & --
    & --
    & --
    & -- & --
    & perception-level
    & \makecell[l]{real traj.\\(offline)}
    & \emph{Real-world seq.} \\
  Follow-Bench\,1.0~\cite{ye2025rpf}
    & --
    & High
    & Mid
    & Back\,+\,Side
    & \cmark & \xmark
    & planning-level
    & \makecell[l]{Social Force Model\\(SFM)~\cite{helbing1995social} / ORCA}
    & 2D Simulator \\
  \textbf{Follow-Bench\,2.0}
    & Mid
    & High
    & High
    & Back\,+\,Side
    & \cmark & \cmark
    & task-level
    & \makecell[l]{NavMesh +\\SFM +\\social activities}
    & Unreal Engine \\
  \bottomrule
  \end{tabular}%
  }
  \begin{minipage}{\linewidth}
    \footnotesize
    \raggedright
    ``--'' denotes out-of-scope axes. E2E denotes end-to-end visual or VLA policies. 
    ORCA* follows EVT-Bench's ORCA-based avoidance setting, although penetration may still occur in dense mesh-agent interactions.
  \end{minipage}
  \vspace{-1.5em}
\end{table}

\subsection{Socially-Aware RPF Methods}
\label{sec:related-methods}

Socially-aware RPF~\cite{islam2019person,honig2018toward,francis2025principles,siva2022robot} has traditionally been studied through modular perception--planning pipelines.The perception module estimates the target person's state, typically using single-object tracking (SOT)~\cite{stark,mixformer2023tpami} or a combination of multi-object tracking (MOT) and target ReID~\cite{zhang2022bytetrack,luo2021multiple,ye2024person,kpr}. The planning module then generates collision-free and kinematically feasible robot motions that follow the target from behind or beside while respecting surrounding pedestrians and the user's spatial preferences~\cite{ye2025rpfsearch,sekiguchi2021uncertainty,ferrer2017robot}. These methods make social constraints explicit and remain strong in close-range following, but they are rarely evaluated together with perception errors in a unified closed-loop benchmark.

EVT~\cite{zhong2021advat,zhong2025unrealzoo,zhong2024empowering,ye2026refertrack} studies a closely related problem from the computer-vision perspective. EVT evaluates whether an embodied agent can keep a target object within its field of view, often considering an episode successful when the target remains visible within a prescribed angular and distance range (90$^{\circ}$ and 7.5m). Compared with socially-aware RPF, this objective permits longer-range target keeping and does not require the robot to maintain a close, socially comfortable following formation around a designated person. Early EVT methods combine visual trackers or vision foundation models (VFM)~\cite{zhong2021advat,zhong2024empowering,wu2025vlm} with reinforcement-learning or diffusion-based motion policies~\cite{zhong2021advat,zhong2024empowering,sun2026instance}. Recent end-to-end and VLA-based policies~\cite{zhang2025embodied,zhang2025uni,liu2026comatrack,liu2025trackvla++,wang2025trackvla,wang2026vlingnav,chu2026abot,wang2026hy,tang2026spatialsv} further unify perception and control in a single model, often using expert trajectories collected in simulation and auxiliary supervision from navigation or person-ReID tasks~\cite{wang2025trackvla,wang2026vlingnav,zhang2025embodied,liu2026comatrack,liu2025trackvla++,zeng2026ezreal,zeng2026taric}. These developments motivate a benchmark that evaluates modular and end-to-end policies under the same socially-aware RPF requirements.

\subsection{Benchmarks for Socially-Aware RPF}
\label{sec:related-bench}

Table~\ref{tab:bench-comparison} compares Follow-Bench~2.0 with representative embodied tracking and RPF-related benchmarks along three capability axes: target re-identification, obstacle/occlusion avoidance, and socially-aware following. Existing EVT benchmarks provide valuable testbeds for visual target keeping, but they cover only part of the RPF problem. Gym-UnrealCV~\cite{qiu2017unrealcv} offers photorealistic scenes with visual distractors and lighting variation, but its interaction structure is largely limited to line-of-sight tracking. DAT~\cite{sunopen} studies active tracking from an aerial perspective, where the agent does not share the pedestrian-level traversability and social-space constraints faced by a ground robot. EVT-Bench~\cite{wang2025trackvla} moves closer to embodied RPF by using humanoid avatars, shorter tracking distances, and ORCA-style pedestrian avoidance. However, it remains primarily back-following-oriented and provides limited evaluation of social comfort, side-following, and dense interaction scenarios such as bottlenecks and queues.

RPF-related benchmarks cover complementary aspects of the task, but do not jointly evaluate perception and planning. TPT-Bench~\cite{ye2025tpt} focuses on long-term target-person tracking from real-world egocentric videos collected in crowded environments, making it suitable for evaluating target re-identification and recovery after target absence. However, it is an offline perception benchmark: the evaluated method does not control the robot, and thus the benchmark cannot reveal how perception errors affect downstream motion, safety, or social comfort. Follow-Bench~1.0~\cite{ye2025rpf} targets the opposite side of the problem by evaluating safety- and comfort-oriented RPF planners under diverse layouts and crowd flows. It supports both back- and side-following, but assumes ground-truth target and pedestrian states in a 2D simulator, leaving the perception-to-planning coupling untested.

Follow-Bench~2.0 closes this gap by evaluating socially-aware RPF as an end-to-end closed-loop task in photorealistic 3D environments. Unlike perception-only benchmarks, it measures how target localization, identity switches, occlusions, and temporary target loss affect robot motion. Unlike planning-only benchmarks, it does not assume perfect perception. Unlike EVT benchmarks, it explicitly evaluates close-range following, social-space intrusion, back- and side-following configurations, pedestrian interactions, and motion comfort. This makes Follow-Bench~2.0 a diagnostic benchmark for studying how modular perception--planning pipelines and end-to-end visual policies behave under the coupled challenges of socially-aware robot person following.

\section{Benchmark Design}
\label{sec:benchmark}

\subsection{Task Formulation}
\label{sec:task}

We formulate socially-aware RPF as a closed-loop embodied task. At the first frame of each episode, the target person is specified by a bounding box. This target specification can initialize VFM-based trackers or modular perception pipelines, and can also be converted into a language description for VLA-style policies. The robot must then follow the designated target through a dynamic 3D environment populated by pedestrians and static obstacles. Unlike embodied visual tracking~\cite{zhong2021advat,zhong2025unrealzoo,zhong2024empowering}, where keeping the target inside the camera view may be sufficient, socially-aware RPF further requires the robot to maintain a close and socially comfortable formation while avoiding collisions and disturbing surrounding pedestrians.

We consider two canonical following configurations. In back-following, the robot stays behind the target, which is common in logistics, patrol, and visually guided following. In side-following, the robot moves beside the target, which better supports companionship and human--robot interaction but introduces stronger constraints from lateral clearance, oncoming pedestrians, and bottlenecks. For either configuration, we distinguish hard \emph{safety requirements} from soft \emph{social objectives}. An episode fails upon a collision, long-term target loss, or failure to return near the target by the end of the episode. Temporary target loss and short-lived deviations from the desired formation are permitted if the robot subsequently recovers the target. Social compliance is evaluated separately by how well the robot maintains the desired target-relative distance and bearing, preserves target visibility, and respects the personal space of the target and surrounding pedestrians. These social measures do not otherwise change the binary success outcome. Appendix~\ref{app:metrics} provides the complete metric definitions.

\subsection{Human-Interactive 3D Environments}
\label{sec:scenarios}

These requirements are rarely independent in everyday human environments. Doorways can simultaneously create queuing, target occlusion, and collision pressure; corners couple geometric occlusion with side-following degradation; dense crowds increase both re-identification ambiguity and motion-planning difficulty; and oncoming pedestrians may occupy the lateral gap required for side-following. A benchmark based only on 2D states or offline perception cannot expose how these factors interact in closed loop. Follow-Bench~2.0 therefore instantiates socially-aware RPF in 3D human-populated environments built on Unreal Engine, which provides photorealistic rendering, animated humanoid pedestrians, configurable sensors, and controllable weather conditions. Our environment design targets \emph{controlled exposure to coupled RPF failure modes}, rather than visual diversity alone.

\begin{figure}[t]
  \centering
  \includegraphics[width=\linewidth]{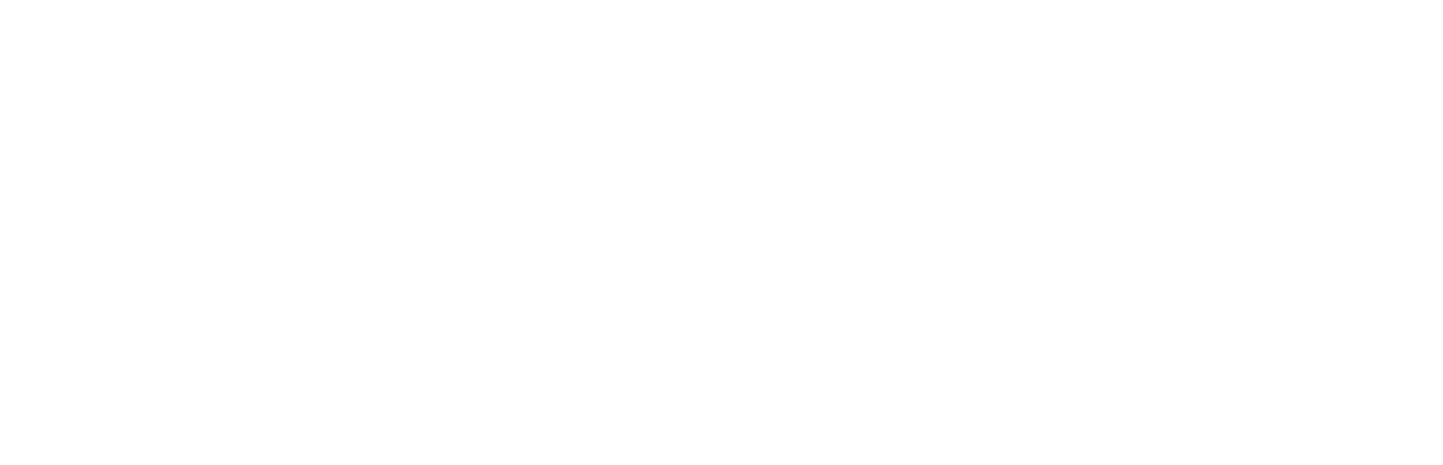}
  \caption{Episode architecture of \textbf{Follow-Bench~2.0} within a human-populated 3D scenario, where the Episode Manager coordinates pedestrians, robot planning, target motion, and evaluation.}
  \label{fig:system-architecture}
  \vspace{-1.5em}
\end{figure}

\paragraph{Episode architecture.}
Each episode consists of a shared scenario layer and four synchronized runtime modules coordinated by an \emph{Episode Manager} (Fig.~\ref{fig:system-architecture}). The scenario layer defines the walkable region, NavMesh~\cite{dosovitskiy2017carla} flow points, structural overlays, NPC-behavior preset, and weather condition. The runtime modules then advance with the simulator clock: an \emph{NPC Controller} drives non-target pedestrians using NavMesh and the Social Force Model (SFM)~\cite{helbing1995social} with scenario-specific behaviors; a \emph{Robot Planner} hosts the evaluated policy and exposes RGB-D, LiDAR, and, for side-following, left/right RGB-D observations; a \emph{Target Person Controller} moves the designated person along reproducible NavMesh routes while keeping it exempt from SFM forces; and an \emph{Evaluation Protocol} logs target visibility, collisions, and social-zone violations using simulator-side privileged signals that are \emph{never} fed back to the policy. Detailed module behavior and sensor specifications are provided in Appendix~\ref{app:scenario-engineering}.

\paragraph{Diagnostic scenarios.}
The four scenarios are designed as \emph{diagnostic probes} rather than merely visually distinct maps. Each scenario stresses a different RPF failure mode while preserving the same evaluation interface. \emph{Open Plaza} tests target re-identification under repeated short pedestrian occlusions in open space. \emph{Street Corner} stresses geometric occlusion and side-following degradation when oncoming pedestrians enter the lateral formation gap. \emph{Cluttered Park} combines distractor orbits, trunk-walk routes, bottleneck zones, and target-look-alike pedestrians, making target identity and social motion planning difficult at the same time. \emph{Narrow Passage} creates queuing, yielding, and short doorway-traversal windows, forcing the robot to balance target keeping against collision avoidance and social-zone intrusion. Per-scenario interaction patterns and instance descriptions are provided in Appendix~\ref{app:scenario-instances}.

\begin{figure}[t]
  \centering
  \includegraphics[width=0.245\linewidth,trim=0 113.33bp 192.83bp 0,clip]{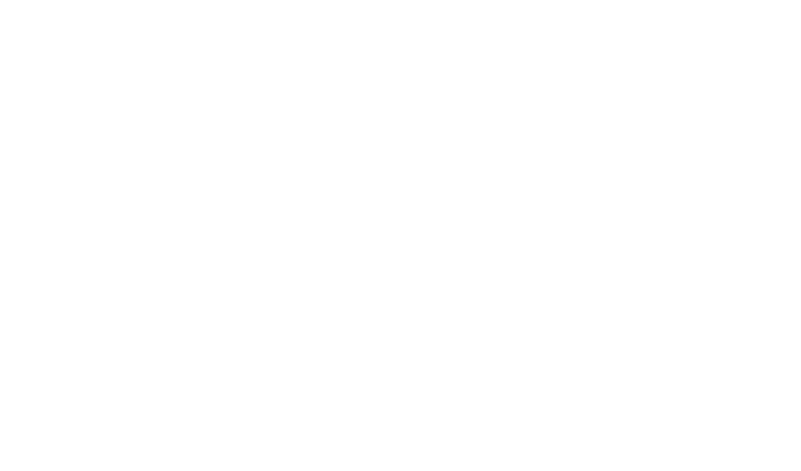}\hfill
  \includegraphics[width=0.245\linewidth,trim=192.83bp 113.33bp 0 0,clip]{figures/optimized/clutter.pdf}\hfill
  \includegraphics[width=0.245\linewidth,trim=0 0 192.83bp 113.33bp,clip]{figures/optimized/clutter.pdf}\hfill
  \includegraphics[width=0.245\linewidth,trim=192.83bp 0 0 113.33bp,clip]{figures/optimized/clutter.pdf}
  \caption{Cluttered Park from \textit{easy} to \textit{legend}. Denser intersecting pedestrian traffic increases target occlusion and lateral-clearance pressure. The target and robot are marked by yellow and red boxes.}
  \label{fig:cluttered-park-difficulty}
  \vspace{-1.2em}
\end{figure}

\paragraph{Composite difficulty preset.}
Each scenario exposes a named NPC-behavior preset, $\textit{difficulty} \in \{\textit{easy},\allowbreak \textit{hard},\allowbreak \textit{boss},\allowbreak \textit{legend}\}$, which controls interaction intensity along four axes: \emph{crowd intensity} (pedestrian count and density), \emph{motion pressure} (walking speed and spawn separation), \emph{social interaction} (crossing, yielding, bottleneck, and shadow-following behaviors), and \emph{identity ambiguity} (target-look-alike pedestrians introduced at \textit{legend}). Weather is independently sampled from $\{\textit{clear}, \textit{cloudy}, \textit{dusk}, \textit{rain}, \textit{fog}\}$, with higher difficulties biased toward lower-visibility conditions. Figure~\ref{fig:cluttered-park-difficulty} illustrates this progression in Cluttered Park. This composite preset is the only task-distribution dimension swept in our main experiments; following position and distance are treated as policy-side configurations in Sec.~\ref{sec:baselines}. Per-scenario numerical settings are reported in Appendix~\ref{app:difficulty-values}, while the underlying knobs for NPC dynamics, pedestrian count, and weather remain individually configurable through the simulator API.

\paragraph{Interaction layer.}
Beyond automated evaluation, Follow-Bench~2.0 includes an interaction layer for policy probing, demonstration collection, and failure diagnosis. Users can manually drive the target person to test policies under customized motion patterns, teleoperate the robot to collect expert demonstrations for imitation learning, and inspect real-time debug signals such as planned trajectories, tracked agents, and detected obstacles. Manual control replaces the corresponding module outputs rather than adding privileged inputs, and the visualizer is read-only with respect to both policy execution and metric computation. Therefore, enabling the interaction layer does not affect formal evaluation episodes. Implementation details and debug-layer screenshots are provided in Appendix~\ref{app:interaction-layer}.

\subsection{Baselines, Evaluation Protocol, and Metrics}
\label{sec:baselines}

\paragraph{Baselines.}
We reproduce $14$ representative policies across modular, VLA-based, and foundation-model-based paradigms under a unified closed-loop protocol. This design allows us to characterize how different systems trade off perception robustness, planning quality, and end-to-end learnability. The complete baseline list is given in Table~\ref{tab:baselines} and detailed in Appendix~\ref{app:baselines}.

For \emph{modular policies}, we adopt the top-performing target-tracking pipeline benchmarked on TPT-Bench~\cite{ye2025tpt}, namely ByteTrack~\cite{zhang2022bytetrack} with Target-ReID~\cite{ye2024person}, and pair it with the three best-performing motion planners from Follow-Bench~1.0~\cite{ye2025rpf}: \textbf{MPC}~\cite{han2023rda}, which follows a designed point; \textbf{MPC w/ Traj.}~\cite{sekiguchi2021uncertainty}, which follows a predicted trajectory; and \textbf{MPC w/ DS.}~\cite{ye2025rpfsearch}, which uses a dynamic search field to handle target loss. By default, the ReID backbone is ResNet18 (\textbf{R18})~\cite{dendorfer2020mot20}. To isolate the effect of perception capacity, we additionally pair MPC w/ Traj. with the heavier \textbf{KPR}~\cite{kpr} backbone. All modular planners share a unified bird's-eye-view (BEV) state lifted from RGB-D and LiDAR observations. As shown in Fig.~\ref{fig:follow-configurations}, back-following uses one front RGB-D view, whereas side-following fuses front, left, and right views. Full pipeline details are provided in Appendix~\ref{app:perception-pipeline}.

For \emph{VLA-based policies}, we evaluate \textbf{TrackVLA}~\cite{wang2025trackvla}, re-trained with a Qwen3-4B~\cite{yang2025qwen3} language backbone. For \emph{foundation-model-based active tracking}, we evaluate \textbf{OA-VAT}~\cite{sun2026instance}, the current state of the art on Gym-UnrealCV~\cite{qiu2017unrealcv} and DAT~\cite{sunopen}. Modular policies are evaluated in both back- and side-following at a desired distance of $1.5$\,m, while TrackVLA and OA-VAT only expose a back-following interface.

\paragraph{Evaluation protocol.}
Each episode starts by specifying the target person with a first-frame bounding box for modular and foundation-model-based policies, or with a language description for VLA policies. The target is initialized as the pedestrian closest to the robot among all candidates. For each scenario--difficulty cell, all baselines share the same $50$ pre-generated episode seeds, enabling paired evaluation under matched actor assignments, start--goal configurations, and weather conditions (Appendix~\ref{app:episode-randomization}). We sweep the composite difficulty preset introduced in Sec.~\ref{sec:scenarios}: each scenario is evaluated at four difficulty levels, \textit{easy}, \textit{hard}, \textit{boss}, and \textit{legend}, where crowd density, weather difficulty, and NPC dynamics are jointly intensified.

Following Follow-Bench~1.0~\cite{ye2025rpf} and EVT-Bench~\cite{wang2025trackvla}, each trial runs for a fixed simulation horizon derived from the target trajectory. Target loss is detected when the robot--target distance exceeds $3$\,m and becomes long-term when this condition persists for more than $N_{\text{lost}}$ consecutive steps. A trial is successful if it contains no collision or long-term target loss and the robot--target distance is at most $3$\,m at termination; shorter losses and temporary formation changes are allowed. Complete metric definitions are provided in Appendix~\ref{app:metrics}.

\begin{center}
  \begin{minipage}{\linewidth}
  \centering
  \includegraphics[width=0.48\linewidth]{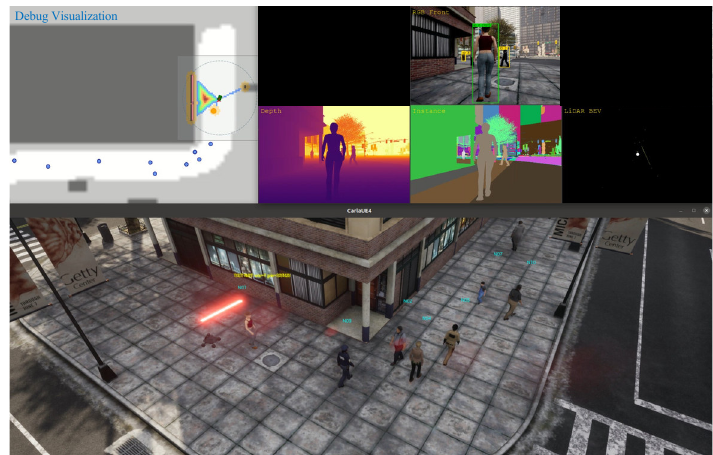}\hfill
  \includegraphics[width=0.48\linewidth]{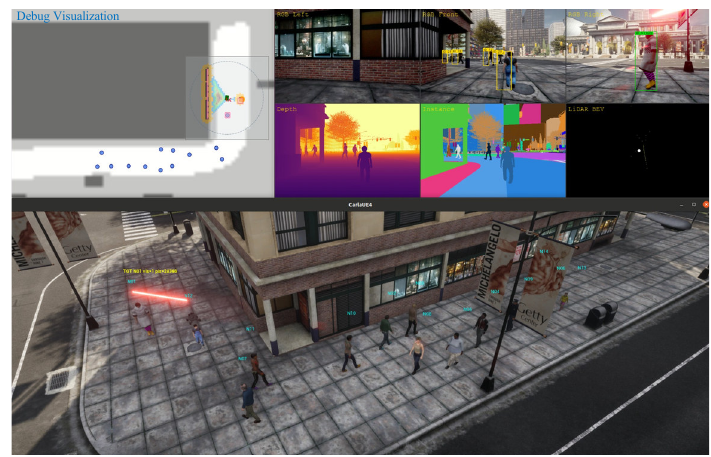}
  \captionsetup{hypcap=false}
  \captionof{figure}{Back-following (left) uses one front RGB-D camera; side-following (right) fuses front, left, and right RGB-D views into the shared BEV interface.}
  \label{fig:follow-configurations}
  \end{minipage}
  \vspace{-1.2em}
\end{center}

\paragraph{Evaluation metrics.}
\phantomsection\label{sec:metrics}
We inherit the safety- and comfort-oriented metric suite from Follow-Bench~1.0~\cite{ye2025rpf}. This is important because socially-aware RPF cannot be characterized by tracking success or collision avoidance alone: a robot may keep the target visible while invading pedestrians' personal space, or avoid collisions while maintaining an uncomfortable formation with the target. Safety metrics measure collision avoidance and recovery from temporary target loss. Social and formation metrics quantify target visibility, desired target-relative distance and bearing, and intrusion into surrounding pedestrians' private zones, while motion-quality metrics quantify smoothness and efficiency.

In the main paper, we report the most discriminative subset: \textbf{SR} (success rate) and \textbf{ASR} (collision-avoidance success rate) for safety; \textbf{TinTPerson} (time ratio inside the target's personal zone, $0.45$--$1.2$\,m) and \textbf{TinPrivate} (time ratio inside surrounding pedestrians' private zone, $<0.45$\,m) for social/formation comfort; and \textbf{Jerk} (motion smoothness measured by the derivative of acceleration) for motion/control quality. The released evaluation harness also logs the full metric suite, including SPL, angle deviation, path length, and average velocity, following Follow-Bench~1.0. Complete metric definitions are provided in Appendix~\ref{app:metrics}.
\section{Evaluation on Follow-Bench~2.0}
\label{sec:experiments}

\subsection{Overall Benchmark Results}
\label{sec:overall}

All policies are evaluated using the same set of $50$ pre-generated episode seeds for each scenario--difficulty cell. Implementation details, including hardware, control rate, and the fixed simulation step are reported in Appendix~\ref{app:impl-details}. Table~\ref{tab:main-results} summarizes back-following performance averaged over the four scenarios and four difficulty presets. We report five safety- and comfort-oriented metrics inherited from Follow-Bench~1.0~\cite{ye2025rpf}, together with per-step observation-to-action latency measured on the evaluation laptop. Per-scenario results are provided in Table~\ref{tab:full-results-all-difficulties} in Appendix; difficulty sensitivity and side-following are analyzed in Sec.~\ref{sec:stress}.

\paragraph{Modular RPF pipelines remain substantially stronger than EVT-style baselines on socially-aware following.}
Every modular configuration in Table~\ref{tab:main-results} outperforms both EVT-style baselines on overall success rate. The strongest modular configuration, MPC w/ Traj.\ + R18, reaches $51.6\%$ SR, whereas TrackVLA and OA-VAT reach only $14.6\%$ and $14.2\%$, respectively. Even the weakest modular configuration, MPC w/ DS.\ + R18, remains about $24$ percentage points ahead of both EVT-style baselines. The gap is more pronounced on TinTPerson: MPC w/ Traj.\ + R18 achieves $92.8\%$, while TrackVLA and OA-VAT achieve only $15.9\%$ and $9.4\%$. This metric directly reflects whether the robot maintains a close and socially comfortable formation with the target.

This gap reveals a mismatch between the EVT objective and socially-aware RPF. TrackVLA is an end-to-end VLA policy trained for visual distractor recovery, while OA-VAT combines a vision-foundation-model tracker with a PID controller. Despite their architectural differences, both are primarily designed to keep the target visible from a relatively safe distance. They do not explicitly optimize close-formation following, pedestrian interaction, or social-zone constraints. In contrast, modular RPF pipelines combine target appearance memory from MOT + ReID with planners that jointly encode desired distance, bearing, collision avoidance, and social-space constraints. OA-VAT provides a clear diagnostic example: it obtains the highest ASR ($78.2\%$), but the lowest SR ($14.2\%$). Its larger effective following distance reduces hard collisions, yet makes the target easier to lose under occlusion or camera-frustum exit. This demonstrates that visibility-oriented active tracking does not directly translate into socially-aware RPF.

\paragraph{Tracking a predicted target trajectory is better than tracking a single offset point.}
With the same R18 perception backbone, MPC w/ Traj.\ + R18 consistently outperforms vanilla MPC + R18 on closed-loop metrics: SR improves from $45.2\%$ to $51.6\%$, ASR from $52.3\%$ to $68.1\%$, and TinTPerson from $51.8\%$ to $92.8\%$. Since the perception module is fixed, the improvement is attributable to the planner. Vanilla MPC tracks a single offset point relative to the target. When nearby pedestrians occupy this set point, the planner must choose between preserving the desired formation and avoiding collisions. MPC w/ Traj.\ instead follows a short-horizon predicted target trajectory, which gives the robot more flexibility to detour around obstacles while still converging back to the target's future path. The large TinTPerson gain indicates that trajectory-level following provides more stable re-convergence after local avoidance. This reversal from the ground-truth-perception ordering in Follow-Bench~1.0, where vanilla MPC performed best, highlights the perception--planning coupling that Follow-Bench~2.0 is designed to expose.

\begin{table}[t]
  \centering
  \caption{Overall benchmark results on Follow-Bench~2.0 under back-following at desired distance $1.5$\,m, averaged over the four scenarios and four difficulty presets ($50$ trials per cell).}
  \label{tab:main-results}
  \footnotesize
  \setlength{\tabcolsep}{3.5pt}
  \renewcommand{\arraystretch}{1.15}
  \resizebox{\linewidth}{!}{%
  \begin{tabular}{l cc cc c c}
    \toprule
    \textbf{Policy}
    & \textbf{SR$\uparrow$}
    & \textbf{ASR$\uparrow$}
    & \textbf{TinTPerson$\uparrow$}
    & \textbf{TinPrivate$\downarrow$}
    & \textbf{Jerk$\downarrow$}
    & \textbf{Latency$\downarrow$} \\
    & (\%) & (\%) & (\%) & (\%) & (m/s$^3$) & (ms) \\
    \midrule
    MPC w/ Traj.\ + R18~\cite{dendorfer2020mot20}
      & \textbf{51.6} & 68.1          & \textbf{92.8} & 11.0          & 37.1          & 42 \\
    MPC w/ Traj.~\cite{sekiguchi2021uncertainty} + KPR~\cite{kpr}
      & 50.8          & 64.2          & 89.5          & 10.6          & 40.6          & 76 \\
    MPC w/ DS.~\cite{ye2025rpfsearch} + R18
      & 38.8          & 71.6          & 35.5          & \textbf{4.3}  & 85.4          & 84 \\
    MPC~\cite{han2023rda} + R18
      & 45.2          & 52.3          & 51.8          & 12.5          & \textbf{18.3} & 38 \\
    TrackVLA~\cite{wang2025trackvla}
      & 14.6          & 64.5          & 15.9          & 9.2           & 45.4          & 64 \\
    OA-VAT~\cite{sun2026instance}
      & 14.2          & \textbf{78.2} & 9.4           & 7.2           & 26.8          & \textbf{21} \\
    \bottomrule
  \end{tabular}%
  }
  \vspace{-1.5em}
\end{table}

\paragraph{A stronger ReID backbone does not necessarily improve closed-loop RPF.}
Holding the planner fixed at MPC w/ Traj., replacing R18 with the heavier KPR backbone brings no closed-loop gain. SR slightly decreases from $51.6\%$ to $50.8\%$, ASR decreases from $68.1\%$ to $64.2\%$, TinTPerson decreases from $92.8\%$ to $89.5\%$, and latency increases from $42$\,ms to $76$\,ms. This suggests that stronger appearance discrimination alone is not sufficient for closed-loop RPF. In difficult perception cases, such as adverse weather, motion blur, and partial human--human occlusion, both R18 and KPR can fail to re-acquire the target. In easier cases where the target is re-detected, the robot may already be in a dense pedestrian flow or narrow passage, where recovery depends more on smooth motion control and crowd-aware planning than on ReID accuracy alone. The remaining headroom therefore lies in degradation-robust perception and smoother socially-aware controllers, rather than in heavier ReID backbones alone.

\paragraph{Search-aware avoidance improves safety but can hurt target recovery.}
MPC w/ DS.\ + R18 achieves the best ASR among modular methods ($71.6\%$) and the lowest TinPrivate overall ($4.3\%$), but it has the lowest modular SR ($38.8\%$). This reflects a trade-off introduced by its search-aware potential field. During target loss, nearby pedestrians act as strong repulsive sources, which reduces collisions and private-zone intrusion. However, the same repulsion can push the robot away from the target's last-seen region. By the time the avoidance maneuver finishes, the target may have left the local search basin and cannot be re-acquired within the remaining horizon. This result shows that socially-aware RPF cannot be solved by independently maximizing avoidance and target keeping. The best-performing configuration, MPC w/ Traj.\ + R18, succeeds because it balances following and avoidance within a single trajectory optimization.

\subsection{Stress Test: Difficulty and Following Configuration}
\label{sec:stress}

\begin{table}[t]
  \centering
  \begin{minipage}[t]{0.47\linewidth}
    \centering
    \caption{Difficulty sensitivity: SR (\%) at each difficulty preset under back-following, averaged over the four scenarios.}
    \label{tab:diff-trend}
    \footnotesize
    \setlength{\tabcolsep}{5pt}
    \renewcommand{\arraystretch}{1.25}
    \resizebox{\linewidth}{!}{%
    \begin{tabular}{@{}l cccc | c@{}}
      \toprule
      \textbf{Method}
      & \textit{easy}
      & \textit{hard}
      & \textit{boss}
      & \textit{legend}
      & \textbf{Avg.} \\
      \midrule
      MPC w/ Traj.\ + KPR   & \textbf{60.5} & 51.5          & 47.0          & 44.0          & 50.8 \\
      MPC w/ Traj.\ + R18   & 58.8          & \textbf{54.0} & \textbf{47.2} & \textbf{46.5} & \textbf{51.6} \\
      MPC w/ DS.\ + R18     & 45.0          & 38.5          & 38.5          & 33.0          & 38.8 \\
      MPC + R18             & 51.3          & 46.2          & 44.7          & 38.5          & 45.2 \\
      TrackVLA              & 18.0          & 17.0          & 10.1          & 13.5          & 14.6 \\
      OA-VAT                & 19.0          & 17.5          & 9.5           & 11.0          & 14.2 \\
      \bottomrule
    \end{tabular}%
    }
  \end{minipage}\hfill
  \begin{minipage}[t]{0.49\linewidth}
    \centering
    \caption{Back- versus side-following: SR (\%) per scenario, averaged over four difficulties.}
    \label{tab:follow-position}
    \footnotesize
    \setlength{\tabcolsep}{4pt}
    \renewcommand{\arraystretch}{1.25}
    \resizebox{\linewidth}{!}{%
    \begin{tabular}{@{}l c cccc | c@{}}
      \toprule
      \textbf{Method}
      & \textbf{Follow}
      & \textbf{Plaza}
      & \textbf{Corner}
      & \textbf{Park}
      & \textbf{Passage}
      & \textbf{Avg.} \\
      \midrule
      \multirow{2}{*}{\makecell[l]{MPC \\ w/ Traj. + KPR}}
      & back & 23.5          & \textbf{96.5} & 0.0 & 83.0          & 50.8 \\
      & side & 1.5           & 2.5           & 0.0 & 0.0           & 1.0 \\
      \cmidrule(lr){1-7}
      \multirow{2}{*}{\makecell[l]{MPC \\ w/ Traj. + R18}}
      & back & 25.0          & 95.4          & 0.0 & \textbf{86.9} & \textbf{51.6} \\
      & side & \textbf{4.0}  & 1.5           & 0.0 & 0.0           & 1.4 \\
      \cmidrule(lr){1-7}
      \multirow{2}{*}{\makecell[l]{MPC \\ w/ DS. + R18}}
      & back & \textbf{35.0} & 90.0          & 0.0 & 30.0          & 38.8 \\
      & side & 2.0           & \textbf{14.5} & 0.0 & 0.0           & \textbf{4.1} \\
      \cmidrule(lr){1-7}
      \multirow{2}{*}{MPC + R18}
      & back & 16.1          & 93.9          & 0.0 & 71.0          & 45.2 \\
      & side & 0.0           & 9.5           & 0.0 & 0.0           & 2.4 \\
      \bottomrule
    \end{tabular}%
    }
  \end{minipage}
  \vspace{-1.5em}
\end{table}

\paragraph{The difficulty preset provides a meaningful stress axis for capable methods.}
Across the four modular configurations, SR decreases monotonically (or near-monotonically) with difficulty: MPC w/ Traj.\ + R18 drops from $58.8\%$ at \textit{easy} to $46.5\%$ at \textit{legend}, vanilla MPC + R18 from $51.3\%$ to $38.5\%$, with MPC w/ DS.\ + R18 showing a single \textit{hard}/\textit{boss} plateau on the same downward trend. The two EVT-style baselines (TrackVLA $18.0 \to 17.0 \to 10.1 \to 13.5\%$; OA-VAT $19.0 \to 17.5 \to 9.5 \to 11.0\%$) are less monotonic, but both already operate near the floor at \textit{easy}, so their per-bucket SR is dominated by seed-level variation. The visual progression and quantitative results together show that the preset provides a meaningful stress axis for capable methods, while current EVT-style baselines are not yet strong enough for fine-grained discrimination.

\paragraph{Per-scenario results show that interaction complexity, rather than pedestrian count alone, determines difficulty.}
Back-following SR exhibits a strong scenario ordering. Averaged over the four modular configurations, Corner reaches about $94\%$ SR, Passage about $68\%$, Plaza about $25\%$, and Park $0\%$ for all methods. This ordering follows the interaction structure of each scenario. Corner is dominated by relatively regular pedestrian flows, and Passage mainly involves single-file queueing through a doorway; in both cases, target motion remains comparatively predictable. Plaza combines random, perpendicular, and parallel pedestrian flows, causing frequent target-path crossings and less reliable short-horizon prediction. Park is deliberately designed as the hardest scenario: the target moves through variable-width narrow passages, including segments that fit only two people abreast, while surrounding pedestrians create simultaneous occlusion and lateral-clearance constraints. Since the same map can be traversed under privileged ground-truth perception and a deliberately designed expert policy, the uniform $0\%$ result indicates a coupled perception--planning failure rather than an infeasible scenario. ReID repeatedly fails under full occlusion, while the planner simultaneously loses enough lateral clearance to recover the following formation.

\paragraph{Side-following remains an open challenge.}
Side-following performance drops sharply for all modular configurations. Among the $16$ method--scenario cells in Table~\ref{tab:follow-position}, $15$ obtain $\leq 4\%$ SR; the only higher value is MPC w/ DS.\ + R18 in Corner ($14.5\%$). Park and Passage yield $0\%$ SR for all side-following configurations, and the average SR drops by about $35$--$50$ percentage points compared with back-following. TrackVLA and OA-VAT are not included in this table because neither exposes a side-following interface, which itself reflects a limitation of existing EVT-style policies. The collapse is caused by two coupled factors. First, side-following imposes stronger spatial constraints than back-following: the robot must stay close to the target while sharing lateral space with oncoming pedestrians, clutter, corners, and doorway boundaries. Second, current RPF planners prioritize maintaining the desired target-relative formation, which can conflict with the stronger avoidance demands of side-following. However, side-following is important for human-centered robot companionship because it enables the robot to walk abreast with the user. Follow-Bench~2.0 therefore exposes side-following as a challenging but underexplored frontier for both modular RPF systems and future end-to-end visual/VLA policies.
\section{Discussion and Conclusion}
\label{sec:conclusion}

We presented \textbf{Follow-Bench~2.0}, an end-to-end 3D benchmark for socially-aware robot person following under coupled perception--planning challenges. It provides diagnostic 3D scenarios, back- and side-following configurations, modular and end-to-end baselines, and safety--comfort metrics in a unified closed-loop protocol. Our evaluation shows that current modular RPF pipelines substantially outperform EVT-style baselines on close and socially comfortable following, but also reveals that interaction-heavy scenarios and side-following remain far from solved. These findings highlight the gap between visibility-oriented active tracking and socially-aware RPF, and position Follow-Bench~2.0 as a diagnostic platform for future research on perception--planning coupling and human-centered robot following.

\paragraph{Limitations.}
Follow-Bench~2.0 has several limitations. \emph{(i)} The four scenarios are curated for diagnostic depth rather than broad visual and geographic coverage; procedural scene generation remains future work. \emph{(ii)} Although Unreal Engine provides controlled and visually realistic rendering, simulated appearance, sensing artifacts, and pedestrian motion do not fully reproduce the real world. In particular, pedestrians follow NavMesh and the Social Force Model rather than trajectories learned from real human interactions, leaving both visual and behavioral sim-to-real gaps. \emph{(iii)} All experiments are simulation-only. Our conclusions concern relative performance and diagnosed failure modes under a matched, controlled closed-loop protocol; the reported absolute success rates should not be interpreted as real-world deployment performance. Future work will study real-robot transfer and reduce the remaining gap through Gaussian-splatting scene reconstructions and more realistic pedestrian appearance and animation.

\paragraph{Broader impacts.}
RPF is a dual-use capability. While it can support service robots in logistics, guidance, assistance, and companionship, it may also enable unwanted persistent tracking, surveillance, or harassment when combined with strong identity recognition. We mitigate these risks by emphasizing socially-aware metrics, including private-zone intrusion, target personal-zone time, and motion comfort, so that benchmark progress is not defined by tracking performance alone. The release documentation explicitly identifies surveillance and non-consensual person following as harmful uses outside the intended scope of the benchmark, and we clearly document the simulation-only scope of our findings to avoid over-claiming real-world readiness.

\section*{Acknowledgments}
{\raggedright
This work was supported in part by Shenzhen Science and Technology Program (No.\ ZDYJ20251211141000001), in part by Shenzhen Science and Technology Program (No.\ SGDX20240115111759002), in part by Meituan Academy of Robotics Shenzhen, in part by the Shenzhen Association for Science and Technology (No.\ XHXS2025-003), and in part by High level of special funds (G03034K003) from Southern University of Science and Technology, Shenzhen, China.

The authors declare no competing interests.\par
}

\setlength{\bibsep}{5pt}
\bibliographystyle{unsrtnat}
\bibliography{references}

\appendix
\clearpage
\section{Appendix}
\label{appendix:supplementary}


\subsection{Visual Comparison with Existing Benchmarks}
\label{app:benchmark-examples}

\begin{figure}[ht]
  \centering
  \includegraphics[width=\linewidth]{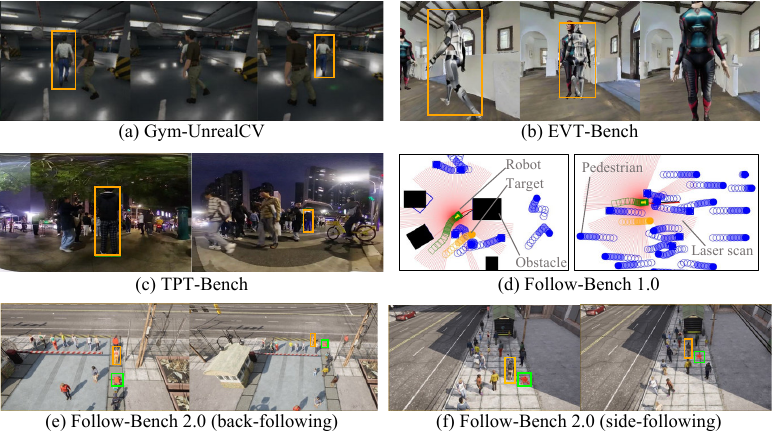}
  \caption{Representative examples from prior EVT/RPF benchmarks and \textbf{Follow-Bench~2.0}. Existing benchmarks capture important yet partial aspects of RPF, such as long-range visual target keeping, offline target-person tracking, or planning with ground-truth states. In contrast, \textbf{Follow-Bench~2.0} evaluates closed-loop RPF in photorealistic 3D scenes, jointly considering target perception, pedestrian-level interactions, back- and side-following, and safety--comfort constraints. In (e), pedestrians queue to pass through a narrow passage. In (f), a side-following robot must avoid oncoming pedestrians and recover the target after occlusion.}
  \label{fig:benchmark-examples}
\end{figure}

Figure~\ref{fig:benchmark-examples} complements the capability-level comparison in Table~\ref{tab:bench-comparison} by showing representative episodes from each benchmark. Existing benchmarks capture important but partial aspects of RPF, including long-range visual target keeping in open scenes (Gym-UnrealCV), back-following with ORCA-based pedestrian avoidance that may still exhibit mesh penetration in dense interactions (EVT-Bench), offline target-person tracking on real-world egocentric videos (TPT-Bench), and 2D motion planning with ground-truth perception (Follow-Bench~1.0). In contrast, Follow-Bench~2.0 evaluates closed-loop RPF in photorealistic 3D scenes with onboard target perception, dense pedestrian-level interactions, both back- and side-following, and a unified safety--comfort metric suite.

\vspace{-1.0em}

\subsection{Baselines (Full List)}
\label{app:baselines}

Table~\ref{tab:baselines} lists every baseline that is implemented and verified under the Follow-Bench~2.0 closed-loop protocol. The main paper restricts the reported numbers to a subset (vanilla MPC, MPC w/ Traj., and MPC w/ DS.\ for the planning module; TrackVLA and OA-VAT for end-to-end paradigms), chosen because these planners give the strongest safety--comfort trade-off on Follow-Bench~1.0 and they jointly let us compare paradigms (modular geometry, predictive modular, end-to-end VLA, foundation-model-based) under a unified protocol. The remaining planners (SFM, DWA, BSO-HFC) and both ReID backbones (ResNet18 and KPR) are fully integrated in the released harness so that benchmark users can run, extend, or replace them without re-implementing the perception--planning interface. Below we briefly describe the working principle of each baseline; the modular planners closely follow the implementations of Follow-Bench~1.0~\cite{ye2025rpf}.

\begin{table}[ht]
  \centering
  \caption{Full baseline list of Follow-Bench~2.0. Modular methods explicitly decouple target tracking from motion planning, while VLA and foundation-model-based policies tightly couple perception and action.}
  \label{tab:baselines}
  \renewcommand{\arraystretch}{1.2}
  \setlength{\tabcolsep}{6pt}
  \footnotesize
  \begin{tabular}{@{}>{\raggedright\arraybackslash}m{0.18\linewidth} >{\raggedright\arraybackslash}m{0.78\linewidth}@{}}
    \toprule
    \textbf{Baselines} & \textbf{Pipeline (tracking $\rightarrow$ planning)} \\
    \midrule
    Modular (back-, side-)
    & ByteTrack~\cite{zhang2022bytetrack} + Target-ReID~\cite{ye2024person} (ResNet18~\cite{dendorfer2020mot20} / KPR~\cite{kpr}) $\rightarrow$ SFM~\cite{ferrer2013robot}, MPC~\cite{han2023rda}, MPC w/ Traj.~\cite{sekiguchi2021uncertainty}, MPC w/ DS~\cite{ye2025rpfsearch}, DWA~\cite{van2022collision}, BSO-HFC~\cite{lyu2025robust}. \\
    \midrule
    VLA (back-)
    & TrackVLA~\cite{wang2025trackvla}: end-to-end RGB + language $\rightarrow$ waypoints. \\
    \midrule
    Foundation-model-based (back-)
    & OA-VAT~\cite{sun2026instance} $\rightarrow$ PID controller. \\
    \bottomrule
  \end{tabular}
\end{table}
\vspace{-1.0em}

\paragraph{Tracking module (modular).} The tracking module first runs ByteTrack~\cite{zhang2022bytetrack} on the front-view RGB stream to obtain per-pedestrian bounding boxes, and then re-identifies the target by comparing each box against a first-frame target template through a Target-ReID network~\cite{ye2024person}; we instantiate the ReID backbone with two capacities, ResNet18~\cite{dendorfer2020mot20} (lightweight, real-time) and KPR~\cite{kpr} (keypoint-aligned, more robust under occlusion and clothing change), so that we can study the latency--accuracy trade-off introduced by the perception backbone.

\paragraph{Motion planners (modular).} The six motion planners receive the BEV state described in Appendix~\ref{app:perception-pipeline} and produce velocity or waypoint commands:
\begin{itemize}\setlength{\itemsep}{1pt}
  \item \textbf{SFM}~\cite{ferrer2013robot,ferrer2017robot}: drives the robot by a sum of social forces --- an attractive force toward the desired following pose behind/beside the target, repulsive forces from each surrounding pedestrian and static obstacle, and pairwise interaction forces that encode social distance preferences.
  \item \textbf{MPC}~\cite{han2023rda}: solves a finite-horizon optimal control problem at each step using the RDA planner~\cite{han2023rda}, whose collision constraints are formulated as a smooth bi-convex dual that tightly tracks the target reference pose while remaining outside the polygonal envelopes of pedestrians and obstacles.
  \item \textbf{MPC w/ Traj.}~\cite{sekiguchi2021uncertainty}: extends vanilla MPC by tracking a reference path generated from the target's Kalman-filter-predicted future trajectory, and modulates the cost weights with the prediction uncertainty so that the robot slows down when the target's trajectory becomes unreliable.
  \item \textbf{MPC w/ DS.}~\cite{ye2025rpfsearch}: extends vanilla MPC with a dynamic search field for safe target re-acquisition after occlusion, switching between an \emph{overtaking} mode (attractive--repulsive cost favoring shorter detours that restore visibility) and a \emph{fluid-following} mode (treating the surrounding crowd as a flow field and following low-density, target-aligned regions) according to overtaking cost.
  \item \textbf{DWA}~\cite{van2022collision}: samples admissible linear and angular velocities inside the dynamic window, forward-simulates each candidate, and selects the trajectory that minimizes a cost combining goal alignment, obstacle clearance, and velocity preference.
  \item \textbf{BSO-HFC}~\cite{lyu2025robust}: hierarchical planner-and-tracker that runs a Hybrid A$^*$ search to obtain a coarse collision-free path, refines it with a uniform B-spline optimizer (penalizing obstacle proximity, dynamic infeasibility, and deviation from a guidance path derived from the target trajectory history), and tracks the optimized trajectory with an MPC controller while an incremental PID adapts the reference speed to maintain the desired following distance.
\end{itemize}

\paragraph{OA-VAT (foundation-model-based).} OA-VAT~\cite{sun2026instance} is a pipeline visual active tracking method that utilizes multi-view feature aggregation via DINOv3~\cite{simeoni2025dinov3} to construct discriminative instance prototypes, effectively preventing confusion from visually similar distractors. After that, a simple PID controller maps target pixel position and apparent size to robot linear/angular velocity for back following, and represents the current state-of-the-art on Gym-UnrealCV~\cite{qiu2017unrealcv} with 93.0\% success rate on average and DAT~\cite{sunopen} with 83.0\% success rate on average.

\paragraph{TrackVLA (VLA).} TrackVLA~\cite{wang2025trackvla} is an end-to-end VLA policy that consumes the front RGB stream and a natural-language target description and directly emits short-horizon waypoints, which a low-level controller converts into velocity commands. None of the recent mainstream VLA models for embodied visual tracking (EVT) release a full training stack: the original TrackVLA has neither open weights nor open training data, and the community-driven OpenTrackVLA project~\cite{opentrackvla2025} only fills this gap partially by re-implementing the inference pipeline and providing a basic training script around a small $0.6$B-parameter checkpoint without the underlying training corpus. To obtain a fair and reproducible VLA baseline at a scale comparable to the original TrackVLA, we therefore rebuild the model on top of OpenTrackVLA. We replace the original $0.6$B language model with the open-source \textbf{Qwen3-4B}~\cite{yang2025qwen3} backbone, keep the rest of the architecture (DINO + SigLIP vision encoders, language-conditioned waypoint decoder) consistent with OpenTrackVLA, and refactor the training code to support DeepSpeed-based distributed training, gradient checkpointing, and mixed-precision optimization on multiple GPUs.

For training data, we re-derive an expert-following corpus on top of EVT-Bench scenes by pairing the A$^*$ algorithm with a variable-speed PID controller, yielding \textbf{1.3M} expert following trajectories with synchronized RGB observations and natural-language target descriptions. To prevent the model from collapsing into a pure action-prediction head and losing person-discrimination ability, we additionally synthesize \textbf{1.3M} ReID-style video question--answering samples on top of SYNTH-PEDES~\cite{zuo2024plip} and \emph{co-train} the model on the union of the two corpora with a balanced sampler. We evaluate the resulting model on the EVT-Bench Distractor (DT) split~\cite{wang2025trackvla} as a parity check: the original TrackVLA with a Vicuna-7B language backbone reports $57.6$\% SR and $63.2$\% TR, while our reproduction with Qwen3-4B reaches $50.6$\% SR and $69.4$\% TR --- slightly lower in success rate but higher in tracking ratio, which we attribute to the smaller backbone together with our delibrately-collected expert trajectories. We treat these numbers as a sufficient parity check to use this model as the TrackVLA baseline throughout the main experiments.

\subsection{Perception-to-Planning Pipeline for Modular Policies}
\label{app:perception-pipeline}

This appendix expands the brief description in Sec.~\ref{sec:baselines}. The modular planners we evaluate all consume a unified bird's-eye-view (BEV) state describing dynamic agents and static obstacles, regardless of how many cameras are used. Each onboard view contributes the same four channels --- the RGB stream (annotated by ByteTrack + Target-ReID for visualisation), the aligned depth map (used to lift each detection to BEV), the instance-segmentation mask (used \emph{only} for offline visibility-GT logging, never as a policy input), and the LiDAR scan (collapsed to a 2D obstacle scan) --- and the difference between back- and side-following is simply how many such views the robot mounts. Figure~\ref{fig:follow-configurations} visualises the two configurations in the main paper.

\paragraph{Back-following: single-view BEV state.} Under back-following, only the front RGB-D camera is mounted (Figure~\ref{fig:follow-configurations}, left). The ByteTrack~\cite{zhang2022bytetrack} + Target-ReID~\cite{ye2024person} stack returns a set of per-pedestrian and target bounding boxes on the front RGB image. For each box, we crop the aligned depth map and mean-filter the depth values inside the box (with simple outlier rejection on the tails) to obtain a single robust depth, and back-project that pixel to a 3D point using the camera intrinsics/extrinsics. Projecting onto the robot's ground plane yields a per-track BEV position. To absorb sensor noise (depth quantization, occasional ReID dropouts, jitter at long range), we maintain a lightweight Kalman filter per track in BEV: the state is position-and-velocity, missing measurements simply propagate the prior, and the smoothed posterior is what the planner observes.

\paragraph{Side-following: cross-view fusion.} Under side-following, the front, left, and right RGB-D cameras are all mounted (Figure~\ref{fig:follow-configurations}, right); the same per-view pipeline is run independently on each, producing three sets of BEV tracks. We then perform a simple cross-view fusion: in the field-of-view overlap region between adjacent cameras, two tracks whose BEV positions fall within a fixed gating distance are treated as the same physical pedestrian and merged by \emph{retaining the older track ID} (and discarding the newer duplicate). This rule keeps pedestrian identities stable as a target or distractor crosses between camera views, without introducing a heavier multi-view tracker.

\paragraph{Static obstacles.} Static obstacles are extracted in the same way as Follow-Bench~1.0~\cite{ye2025rpf}: the 3D LiDAR scan is collapsed onto the robot's ground plane to a virtual 2D scan, range returns are clustered, and each cluster is fit to a geometric obstacle polygon. The resulting polygons are concatenated with the BEV pedestrian tracks above and handed to the downstream motion planner.

\paragraph{Localization assumption.} For simplicity, the robot's global pose is assumed to be available during evaluation; this isolates the perception/planning failure modes that Follow-Bench~2.0 is designed to expose from SLAM-induced drift. In real deployments, the pose would be supplied by an external module such as LiDAR-inertial SLAM, which is orthogonal to the perception-to-planning interface studied here.

\subsection{Evaluation Metrics (Full List)}
\label{app:metrics}

Table~\ref{tab:metrics} lists the full metric suite inherited from Follow-Bench~1.0~\cite{ye2025rpf} and EVT-Bench~\cite{wang2025trackvla}, and logged by the released harness for every closed-loop trial. We separate hard safety requirements, which determine binary episode success, from soft social/formation and motion-quality measures. The main paper reports the most discriminative subset (SR, ASR, TinTPerson, TinPrivate, and Jerk); the released logs support the complete suite without rerunning the simulator.

\paragraph{Success and target-loss events.}
Let $d^{c}_{RT}(k)=\|\mathbf{p}^{R}_{k}-\mathbf{p}^{T}_{k}\|$ be the center-to-center robot--target distance. Target loss is detected when $d^{c}_{RT}(k)>3$\,m and is considered long-term when it persists for more than $N_{\text{lost}}$ consecutive steps. An episode succeeds if no collision or long-term target loss occurs and $d^{c}_{RT}(N)\leq3$\,m at termination. Thus, temporary target loss and temporary deviation from the desired back- or side-following formation are allowed, but the robot must recover and finish near the target. Separately, a visual-loss event begins when the target leaves the policy's observable views and is successfully recovered if the correct target is re-identified within $N_{\text{lost}}$ steps; these events define SSR and SPL.

\paragraph{Per-step quantities.}
For an episode containing $N$ valid steps at interval $\Delta t$, let $\mathbf{p}^{R}_{k}$ and $\mathbf{v}^{R}_{k}$ denote the executed robot position and linear velocity, $z_k\in\{0,1\}$ denote simulator-side target visibility, and
$d_{RH}(k)=\|\mathbf{p}^{R}_{k}-\mathbf{p}^{H}_{k}\|-(R_r+R_h)$ denote robot--human surface clearance. Acceleration and jerk are computed from the executed velocities as $\mathbf{a}_k=(\mathbf{v}^{R}_{k}-\mathbf{v}^{R}_{k-1})/\Delta t$ and $\mathbf{j}_k=(\mathbf{a}_k-\mathbf{a}_{k-1})/\Delta t$. Metrics are first computed over each episode's valid steps or events and then averaged across evaluation episodes.

\begin{table}[t]
  \centering
  \caption{Full evaluation metric suite of Follow-Bench~2.0, inherited from Follow-Bench~1.0~\cite{ye2025rpf}. Hard safety metrics determine or diagnose episode success; social/formation and motion-quality metrics are soft measures.}
  \label{tab:metrics}
  \renewcommand{\arraystretch}{1.2}
  \setlength{\tabcolsep}{5pt}
  \footnotesize
  \begin{tabular}{@{}>{\raggedright\arraybackslash}m{0.18\linewidth} >{\raggedright\arraybackslash}m{0.78\linewidth}@{}}
    \toprule
    \textbf{Category} & \textbf{Metric (unit, $\uparrow$/$\downarrow$): description} \\
    \midrule
    \multirow{4}{*}{\textbf{Safety}}
    & Success Rate (SR) (\%, $\uparrow$): fraction of episodes with no collision or long-term target loss and a terminal robot--target distance of at most $3$\,m. \\
    & Avoidance Success Rate (ASR) (\%, $\uparrow$): fraction of episodes completed without a collision. \\
    & Search Success Rate (SSR) (\%, $\uparrow$): fraction of visual-loss events in which the correct target is re-identified within $N_{\text{lost}}$ steps. \\
    & SPL ($\uparrow$): mean $S_jL_j/\max(L_j,P_j)$ over visual-loss events $j$, where $S_j$ indicates successful recovery and $L_j$, $P_j$ are the shortest feasible and executed recovery-path lengths. \\
    \midrule
    \multirow{4}{*}{\makecell[l]{\textbf{Social /}\\\textbf{formation}}}
    & Target Visibility Ratio (TVR) (\%, $\uparrow$): $\frac{1}{N}\sum_k z_k$, the fraction of valid steps in which the target is visible. \\
    & TinTPerson (\%, $\uparrow$): $\frac{1}{N}\sum_k\mathbb{1}[0.45\leq d_{RT}(k)\leq1.2\text{ m}]$, the time ratio inside the target's personal zone. \\
    & TinPrivate (\%, $\downarrow$): fraction of valid steps for which $d_{RH_i}(k)<0.45$\,m for at least one surrounding pedestrian $i$. \\
    & Angle Deviation (deg, $\downarrow$): mean wrapped absolute deviation between the actual target-relative bearing and the configured back- or side-following bearing. \\
    \midrule
    \multirow{3}{*}{\makecell[l]{\textbf{Motion /}\\\textbf{control}}}
    & Jerk (m/s$^3$, $\downarrow$): $\frac{1}{N}\sum_k\|\mathbf{j}_k\|$, with unavailable boundary samples set to zero. \\
    & Path Length (m, $\downarrow$): $\sum_{k=2}^{N}\|\mathbf{p}^{R}_{k}-\mathbf{p}^{R}_{k-1}\|$. \\
    & Avg. Velocity (m/s): $\frac{1}{N}\sum_k\|\mathbf{v}^{R}_{k}\|$. \\
    \bottomrule
  \end{tabular}
  \vspace{-1.0em}
\end{table}

\subsection{Human-Interactive 3D Environments: Implementation Details}
\label{app:scenario-engineering}

This appendix expands Sec.~\ref{sec:scenarios} with the implementation details of the human-interactive 3D environment layer. Sec.~\ref{sec:scenarios} focuses on the diagnostic role of each scenario and the high-level episode architecture; here we specify the shared scenario substrate, episode randomization, episode-loop synchronisation, per-module behaviour, the interaction layer, per-scenario diagnostic instances, per-scenario difficulty instantiations, and the offline logging / re-evaluation pipeline.

\subsubsection{Shared Scenario Substrate}
\label{app:scenario-substrate}

Each scenario is authored from a small set of shared components. A walkable region of interest defines where pedestrians and robots may move. Flow points anchor target routes and pedestrian streams on the NavMesh. Structural overlays optionally define queue stations, bottleneck zones, or hand-annotated branch routes. A named NPC-behavior preset then selects the pedestrian parameters for the current difficulty level.

The walkable region is stored as an occupancy-style grid with free-space, ROI, and raw-obstacle masks. Flow points are stored as world-frame waypoints. For scenarios that require additional structure, Narrow Passage uses fixed queue and passage routes, while Cluttered Park uses bottleneck zones and trunk-walk routes. This substrate keeps scenario authoring separate from policy evaluation: adding a new scenario requires annotating the walkable region, flow points, and optional overlays, while the episode loop, policy interface, and metrics remain unchanged.

\subsubsection{Distribution and Customization Workflow}
\label{app:customization}

The simulator is distributed as a pre-compiled Unreal Engine executable, so benchmark users can launch the provided environments without compiling the engine or project source. A Python client API wraps simulator initialization, episode reset, sensing, and robot control. It exposes synchronized RGB and depth images, semantic/instance segmentation for development and evaluation, and LiDAR point clouds, while accepting robot velocity commands or short-horizon waypoints through the shared control interface. Simulator-side privileged segmentation used for formal metric computation remains excluded from policy observations. The Python evaluation and baseline packages retain their normal software dependencies, which are specified in the release documentation.

Adding a new environment follows a scene-to-scenario workflow. A user first navigates the environment in a third-person spectator view and selects a region of interest. The scene-construction script converts this region into an occupancy-style map containing free-space, ROI, and obstacle masks. The user then annotates flow points and, when needed, queue stations, bottleneck zones, or branch routes. On top of these annotations, the scenario scripts configure target and pedestrian appearances, sample valid start--goal pairs, construct NavMesh routes, and set pedestrian behaviors and weather. The same pipeline subsequently launches policies, records synchronized trajectories and events, computes metrics offline, and generates debug visualizations. New interaction patterns can therefore be introduced through configuration and Python scripts without modifying or recompiling the Unreal Engine executable.

\subsubsection{Episode Randomization and Paired Replay}
\label{app:episode-randomization}

Each scenario--difficulty cell is represented by a fixed manifest of $50$ episode seeds. When an episode is generated, its seed samples the target appearance from a database of $52$ humanoid actors and instantiates surrounding pedestrians from the remaining actors. Target and pedestrian start--goal pairs are sampled from predefined valid regions of the scene, subject to the scenario's walkability and spawn-separation constraints. The seed also fixes the robot initialization, weather sample, NPC-behavior preset, spawn order, and all scenario-specific random choices. Consequently, actor appearances and start--goal configurations vary across seeds even when the scenario and difficulty are fixed.

The resulting episode specification is stored and reused by every evaluated policy. In particular, a sampled target start--goal pair is converted into a NavMesh route once and then replayed deterministically across policies; weather, initial states, actor assignments, and NPC random draws are likewise matched. Non-target pedestrians remain reactive to the robot through the Social Force Model, so their realized trajectories may diverge after policies produce different robot motions. This divergence is part of the closed-loop interaction being evaluated, while the exogenous episode randomization remains paired and reproducible.

\subsubsection{Episode Loop and Synchronisation}
\label{app:episode-loop}

The Episode Manager advances the simulator and all modules synchronously. At each control tick, it fetches robot sensor data, builds the policy-side observation, calls the policy, applies the returned action, advances the simulator, and records the synchronized state. Both the simulator and policy operate at $\Delta t=0.1$\,s (10\,Hz). The simulator advances only after the current policy action has been returned, so every evaluated trajectory is sampled at the same simulated-time interval even when wall-clock inference latency differs across policies.

Policy-side observations contain only the sensors and target specification permitted for the corresponding baseline. Modular policies receive RGB-D and LiDAR measurements processed by the onboard perception pipeline into estimated BEV tracks and obstacle geometry; visual and VLA policies receive their specified RGB observations and target cue. Optional left/right RGB-D views are additionally available for side-following. Ground-truth target and pedestrian states, the simulator-side instance-segmentation image, target-visibility flag, target pixel count, contact events, and social-zone annotations are evaluator-only privileged signals: they are logged for metric computation but never exposed to an evaluated policy. Policy actions are represented as linear and angular velocity commands; policies that produce waypoints (e.g., TrackVLA) are wrapped by an adapter that converts them into the same low-level action interface.

\begin{table}[t]
  \centering
  \caption{Diagnostic roles of the four human-interactive scenarios in Follow-Bench~2.0. Each scenario is designed around a different source of socially-aware RPF failure rather than serving as a visually different map.}
  \label{tab:scenario-diagnosis}
  \renewcommand{\arraystretch}{1.18}
  \setlength{\tabcolsep}{5pt}
  \footnotesize
  \begin{tabular}{@{}p{0.13\linewidth} p{0.25\linewidth} p{0.28\linewidth} p{0.25\linewidth}@{}}
    \toprule
    \textbf{Scenario}
    & \textbf{Human-interaction pattern}
    & \textbf{Diagnosed RPF failure mode}
    & \textbf{Why 3D closed-loop evaluation matters} \\
    \midrule
    Open Plaza
    & Bidirectional pedestrian streams cross the target's route in open space.
    & Short repeated occlusions, target ReID ambiguity, and formation drift under passing pedestrians.
    & The robot must recover from intermittent visual evidence while its own motion changes future viewpoints and occlusions. \\
    Street Corner
    & Counter-flow pedestrians converge near a corridor junction and occupy the lateral gap around the target.
    & Geometric target occlusion, side-following degradation, and close-passing conflicts.
    & The failure emerges from the coupling between corner geometry, robot viewpoint, pedestrian flow, and lateral formation constraints. \\
    Cluttered Park
    & Distractor orbits, directed flows, trunk-walk routes, bottlenecks, and a shadow follower create dense local interactions.
    & Appearance-based target ReID failure, near-target view contesting, local congestion, and socially uncomfortable motion.
    & The robot must distinguish the target from visually similar pedestrians while planning through dynamic crowd structure. \\
    Narrow Passage
    & Pedestrians queue, yield, and pass through a doorway-like bottleneck under a concurrency limit.
    & Queue negotiation, doorway-following failure, pedestrian collision, target loss, and social-zone intrusion.
    & The robot must decide when to wait, yield, or follow the target through a narrow passage without disrupting the queue. \\
    \bottomrule
  \end{tabular}
\end{table}

\subsubsection{Module Details}
\label{app:module-details}

The four modules and the cross-cutting interaction layer outlined in Sec.~\ref{sec:scenarios} (Figure~\ref{fig:system-architecture}) are detailed below.

\paragraph{Pedestrian module.} The pedestrian module uses NavMesh-based global navigation~\cite{dosovitskiy2017carla} and Social Force Model (SFM) local interactions~\cite{helbing1995social}, with scenario-specific behaviours layered on top. Open Plaza uses paired counter-flow traversal between flow points. Street Corner uses grid-based routing over the walkable region so that opposing flows converge near the junction. Cluttered Park mixes distractor orbits, directed corridor flows, trunk-walk routes, bottleneck zones, and a shadow follower. Narrow Passage uses a queue-passage state machine (\textsc{Idle}\,$\rightarrow$\,\textsc{Approach}\,$\rightarrow$\,\textsc{Queue}\,$\rightarrow$\,\textsc{Pass}) with a global concurrency limit on the number of agents simultaneously inside the doorway. A long-term rescue mechanism re-routes agents whose accumulated displacement is too small over time, preventing persistent deadlocks.

\paragraph{Robot module.} The default robot platform is Scout-Mini. The sensor suite contains a front RGB-D camera, optional left and right RGB-D cameras for side-following, a single-line LiDAR, and a simulator-side instance-segmentation camera reserved for visibility ground truth (never exposed to the policy). RGB-D cameras default to $800\times600$ resolution and a $90^{\circ}$ field of view; the LiDAR uses a $30$\,m range and returns points in the sensor frame. Each episode writes a calibration file containing camera intrinsics/extrinsics and LiDAR parameters.

\paragraph{Target person module.} For each episode seed, the target's sampled start--goal pair is converted into a fixed NavMesh route that is replayed across policies. The target is exempt from SFM forces so this route remains reproducible under different surrounding-pedestrian interactions. A manual-control hook can replace the scripted route when collecting demonstrations or running stress tests.

\paragraph{Evaluation module.} The evaluation module subscribes to target and pedestrian trajectories, instance-level visibility, contact events, and policy-side commands. These signals are logged but not returned to the policy. Per-tick records are stored with robot, target, human, visibility, planner-debug, and timing fields. At the end of each episode, the stored trajectory is converted into the safety and comfort metrics defined in Appendix~\ref{app:metrics}; because the conversion is offline, stored episodes can be re-evaluated with updated metric thresholds or alternative metric subsets.

\subsubsection{Interaction Layer}
\label{app:interaction-layer}

The interaction layer contains manual-control hooks and a real-time debug visualiser. Manual-control hooks can substitute for the policy output of the robot, the scripted target controller, or selected pedestrian controllers. The debug visualiser renders the walkable region, robot pose, target highlight, pedestrian positions, LiDAR-derived obstacles, planned trajectories, and planner-supplied debug information.

The interaction layer is decoupled from metric computation: manual control replaces the corresponding module output rather than adding an extra input, and the debug visualiser is read-only with respect to both policy and evaluation. Formal evaluation episodes therefore use the same recorded signals regardless of whether visualisation is enabled.

This decoupling turns Follow-Bench~2.0 from a fixed benchmark into an interactive research platform. A user can drive the target along an unusual path, force a pedestrian to block a doorway, reproduce a corner occlusion, create a close-passing event, or teleoperate the robot to collect expert behaviour --- all without changing the metric computation. The same interface supports data-centric policy improvement: a researcher can replay a failure, adjust the target or pedestrian behaviour, test whether the policy recovers, and collect corrective demonstrations for imitation learning or DAgger-style data aggregation. This capability is especially important for VLA-based and foundation-model-based followers, whose performance often depends on the diversity and diagnostic value of training episodes.

\paragraph{Trajectory-data availability.}
The current release focuses on diagnostic evaluation and does not include a trajectory dataset for policy training. We are developing a scalable collection pipeline in which a privileged expert planner generates initial trajectories from ground-truth target, pedestrian, and obstacle states. Teleoperation is then used to provide corrective demonstrations when the expert violates task constraints, following a DAgger-style data-aggregation process. We will release the resulting trajectory dataset once its collection and validation pipeline is mature.

\subsubsection{Diagnostic Roles and Per-Scenario Instances}
\label{app:scenario-instances}

Table~\ref{tab:scenario-diagnosis} expands the compact two-column scenario table in Sec.~\ref{sec:scenarios} with the human-interaction pattern that each scenario implements and the reason 3D closed-loop evaluation is needed to expose its failure mode. Each scenario is then described as a self-contained instance below.

\begin{figure}[!htbp]
  \centering
  \includegraphics[width=\textwidth]{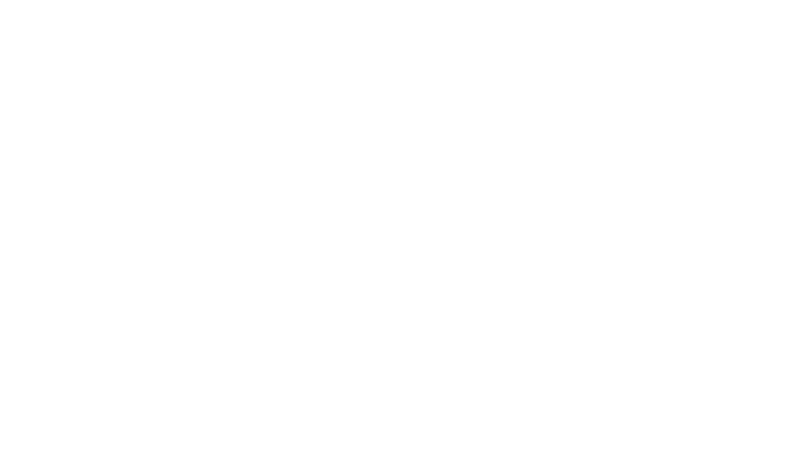}
  \caption{Difficulty progression in the Open Plaza scenario. From \textit{easy} to \textit{legend}, the scene introduces denser bidirectional pedestrian traffic, increasing short-term target occlusion, target re-identification ambiguity, collision risk, and formation-keeping difficulty. The target person and robot are highlighted with yellow and red boxes, respectively.}
  \label{fig:open-plaza-difficulty}
\end{figure}

\paragraph{Open Plaza.}
Open Plaza contains bidirectional pedestrian streams that cross the
target's route in an otherwise open area. The main challenge is not
obstacle avoidance around fixed geometry, but maintaining the target
identity through repeated short occlusions and target-look-alike
distractors. These crossings create episodes where the target is visible,
briefly hidden, and then visible again from a changed robot viewpoint. The
robot must preserve back- or side-following formation through these
interruptions without switching to a visually similar pedestrian. As shown
in Figure~\ref{fig:open-plaza-difficulty}, increasing the difficulty preset
progressively densifies the pedestrian traffic around the target, making
target re-identification, collision avoidance, and formation keeping more
tightly coupled.

\begin{figure}[!htbp]
  \centering
  \includegraphics[width=\textwidth]{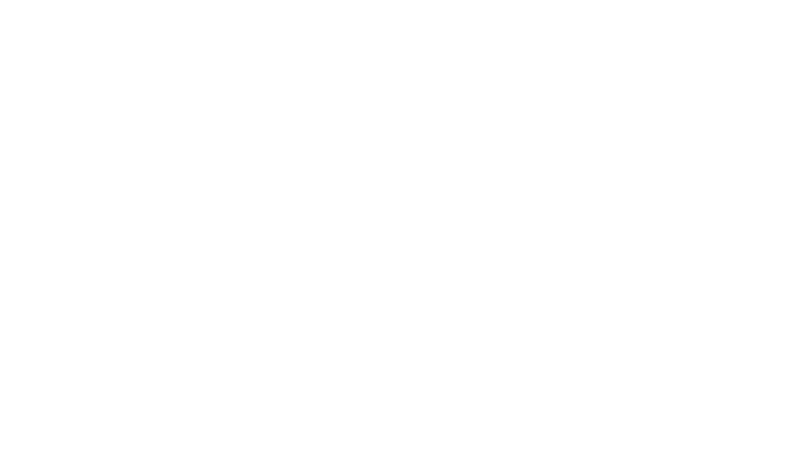}
  \caption{Difficulty progression in the Street Corner scenario. From \textit{easy} to \textit{legend}, denser pedestrian traffic around the corridor junction increases geometric occlusion, close-passing conflicts, collision risk, and side-following difficulty. The target person and robot are highlighted with yellow and red boxes, respectively.}
  \label{fig:street-corner-difficulty}
\end{figure}

\paragraph{Street Corner.}
Street Corner places the target and robot in a corridor-and-junction
layout. Oncoming pedestrians converge near the corner and can occupy the
gap between the robot and the target. The setting is especially
challenging for side-following, where lateral clearance is limited and the
robot must negotiate close encounters without drifting into the target or
nearby pedestrians. Geometric occlusion at the corner further interrupts
target visibility, turning a local planning problem into a coupled
perception--planning challenge. As shown in
Figure~\ref{fig:street-corner-difficulty}, increasing the difficulty preset
progressively densifies pedestrian traffic around the junction, making
target visibility, close-range avoidance, and side-following formation
more tightly coupled.

\paragraph{Cluttered Park.}
Cluttered Park combines multiple pedestrian behaviors in a dense branching
layout. Distractor orbits are slow closed-loop or circular routes repeatedly
traversed by non-target pedestrians near the target, creating persistent local
traffic and visual interference. Unlike these recurring orbits, corridor-flow
and trunk-walk pedestrians traverse their assigned routes once, introducing
transient intersections with the target's path. A shadow follower can further
contest the robot's near-target view. At the highest difficulty,
target-look-alike pedestrians stress
appearance-based target re-identification. Bottleneck zones create local
congestion and partial occlusion as a consequence of pedestrian
interaction rather than scripted visibility masks. As shown in
Figure~\ref{fig:cluttered-park-difficulty}, increasing the difficulty preset
progressively densifies the local crowd around the target and introduces
more intersecting pedestrian motions, making target re-identification,
occlusion recovery, obstacle avoidance, and socially comfortable motion
more tightly coupled.

\begin{figure}[!htbp]
  \centering
  \includegraphics[width=\textwidth]{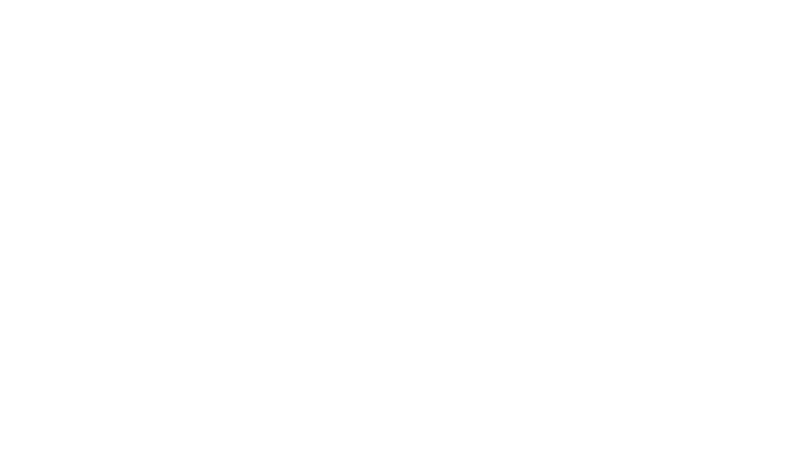}
  \caption{Difficulty progression in the Narrow Passage scenario. From \textit{easy} to \textit{legend}, pedestrian traffic around the passage becomes denser, increasing queueing pressure, target occlusion, collision risk, and social-zone intrusion risk. The target person and robot are highlighted with yellow and red boxes, respectively.}
  \label{fig:narrow-passage-difficulty}
\end{figure}

\paragraph{Narrow Passage.}
Narrow Passage models a doorway-like transition where the robot must
follow the target through a constrained entrance while pedestrians queue
and pass in both directions. The queue-passage state machine creates
waiting, yielding, congestion, and short traversal windows. A successful
policy must decide when to follow the target through the passage while
avoiding collisions and social-zone violations around the queue, directly
testing whether a follower can remain socially compliant when target
keeping, collision avoidance, and queue negotiation conflict. As shown in
Figure~\ref{fig:narrow-passage-difficulty}, increasing the difficulty preset
progressively densifies the pedestrian traffic around the passage, making
target occlusion, collision avoidance, and social-zone compliance more
tightly coupled.

\subsubsection{Per-Scenario Difficulty Instantiations}
\label{app:difficulty-values}

The composite difficulty preset $\textit{difficulty} \in \{\textit{easy}, \textit{hard}, \textit{boss}, \textit{legend}\}$ introduced in Sec.~\ref{sec:scenarios} defines an interaction-intensity gradient rather than a single universal numerical setting. Table~\ref{tab:difficulty} summarises the qualitative progression along four axes that apply to every scenario, and Table~\ref{tab:appendix-difficulty-values} lists the exact per-scenario parameter values at each preset. Weather is independently sampled per episode from $\{\textit{clear}, \textit{cloudy}, \textit{dusk}, \textit{rain}, \textit{fog}\}$ and recorded in the episode metadata; for \emph{boss} and \emph{legend} the sampler is biased toward lower-visibility conditions (\textit{dusk}, \textit{rain}, \textit{fog}) to additionally stress visual perception.

\begin{center}
  \begin{minipage}{\linewidth}
  \centering
  \captionsetup{hypcap=false}
  \captionof{table}{Qualitative progression of the named NPC-behavior presets across the four interaction-intensity axes shared by all four scenarios. Per-scenario parameter values are provided in Table~\ref{tab:appendix-difficulty-values}.}
  \label{tab:difficulty}
  \renewcommand{\arraystretch}{1.12}
  \setlength{\tabcolsep}{4pt}
  \footnotesize
  \resizebox{0.88\linewidth}{!}{%
  \begin{tabular}{@{}l p{0.18\linewidth} p{0.18\linewidth} p{0.18\linewidth} p{0.18\linewidth}@{}}
    \toprule
    \textbf{Preset} & \textbf{Crowd intensity} & \textbf{Motion pressure} & \textbf{Social interaction} & \textbf{Identity ambiguity} \\
    \midrule
    \textit{Easy}   & Sparse pedestrian flow      & Low walking speed and loose spacing   & Mild crossings or queueing                                                  & No target-look-alike pedestrians                       \\
    \textit{Hard}   & Denser pedestrian flow      & Faster motion and reduced spacing     & More frequent crossings, yielding, or bottleneck interactions               & No target-look-alike pedestrians                       \\
    \textit{Boss}   & High-density pedestrian flow & Aggressive motion and tight spacing   & Strong congestion, shadow following, or simultaneous-passage pressure       & No target-look-alike pedestrians                       \\
    \textit{Legend} & Highest scenario-specific density & Highest motion pressure         & Strongest social disturbance and bottleneck pressure                        & Target-look-alike pedestrians are introduced           \\
    \bottomrule
  \end{tabular}%
  }
  \end{minipage}
\end{center}

\begin{center}
  \begin{minipage}{\linewidth}
  \centering
  \captionsetup{hypcap=false}
  \captionof{table}{Per-scenario instantiation of the named NPC-behavior presets used in Follow-Bench~2.0.}
  \label{tab:appendix-difficulty-values}
  \renewcommand{\arraystretch}{1.1}
  \setlength{\tabcolsep}{4pt}
  \scriptsize
  \resizebox{\linewidth}{!}{%
  \begin{tabular}{@{}l l c c c c@{}}
    \toprule
    \textbf{Scenario} & \textbf{Parameter} & \textbf{Easy} & \textbf{Hard} & \textbf{Boss} & \textbf{Legend} \\
    \midrule
    Open Plaza & Walker count                 & 10         & 15         & 20         & 25         \\
    Open Plaza & Crowd speed (m/s)            & 1.20--1.80 & 1.40--2.10 & 1.60--2.40 & 1.80--2.70 \\
    Open Plaza & Target speed (m/s)           & 0.80--1.40 & 0.80--1.40 & 0.80--1.40 & 0.80--1.40 \\
    \midrule
    Street Corner & Walker count                  & 8         & 12        & 16        & 20        \\
    Street Corner & Lateral band half-width (m)   & 4.0       & 3.0       & 2.5       & 2.0       \\
    Street Corner & Min.\ spawn separation (m)    & 2.0       & 1.4       & 1.0       & 0.8       \\
    Street Corner & Speed range (m/s)             & 0.5--1.0  & 0.7--1.3  & 0.9--1.55 & 1.1--1.8  \\
    \midrule
    Cluttered Park & Walker count                     & 35  & 35  & 35  & 35  \\
    Cluttered Park & Corridor flow probability    & 0.3 & 0.7 & 1.0 & 1.0 \\
    Cluttered Park & Shadow-follower probability  & 0.0 & 0.5 & 1.0 & 1.0 \\
    Cluttered Park & Shadow delay (s)             & 5.0 & 5.0 & 3.0 & 1.5 \\
    Cluttered Park & Max pedestrians per segment  & 2   & 3   & 4   & 6   \\
    Cluttered Park & Min.\ gap interval (s)       & 8.0 & 5.0 & 3.0 & 1.5 \\
    Cluttered Park & Trunk-walk max per route     & 1   & 2   & 3   & 4   \\
    Cluttered Park & Trunk dispatch interval (s)  & 12.0 & 8.0 & 5.0 & 3.0 \\
    \midrule
    Narrow Passage & Concurrent in passage        & 2          & 4          & 6          & 8          \\
    Narrow Passage & Speed range (m/s)            & 0.80--1.04 & 0.88--1.20 & 0.96--1.36 & 1.12--1.52 \\
    \bottomrule
  \end{tabular}%
  }
  \end{minipage}
\end{center}

\paragraph{API exposure.} The three sub-knobs (NPC-behaviour parameters, walker count, weather distribution) remain individually exposed in the simulator API; benchmark users who wish to study any one component in isolation can do so without touching the other two.

\FloatBarrier
\subsubsection{Logging and Offline Evaluation}
\label{app:logging-evaluation}

Each trial stores metadata, per-tick states, collision events, and final evaluation results. The metadata records benchmark version, scenario, planner, perception setting, following formation, pedestrian count, target track identifier, trial identifier, control interval, duration, robot and human geometry, and termination settings. The per-tick log records robot, target, human, visibility, planner-debug, and timing information. Collision events are stored separately with collision type and participants. Offline evaluation reads the stored episode and recomputes the metric suite without running the simulator, which lets benchmark users add new metric definitions without rerunning the closed-loop trials.

\subsection{Platform Assets and Robot Embodiments}
\label{app:platform-assets}

\begin{figure}[!htbp]
  \centering
  \includegraphics[width=0.85\linewidth]{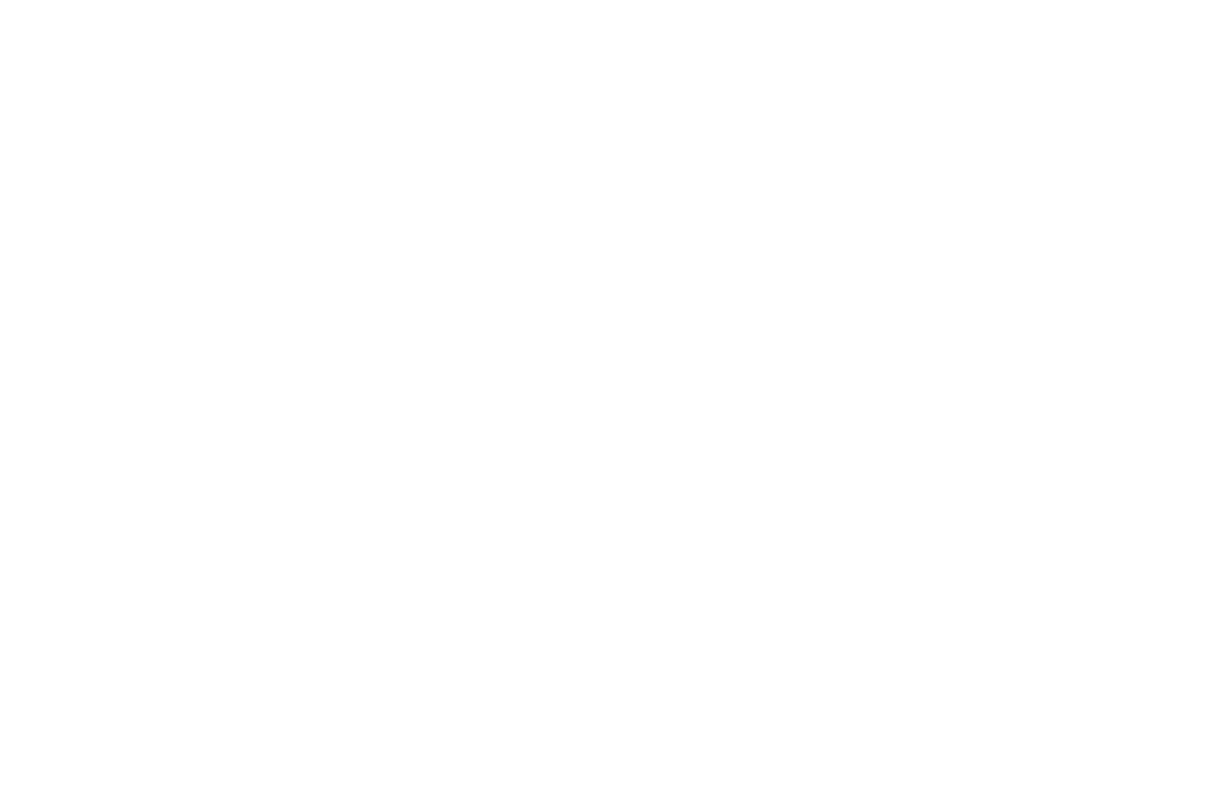}
  \caption{Overview of robot assets in Follow-Bench~2.0. The benchmark includes multiple robot embodiments --- wheeled, quadruped, and humanoid --- allowing policies to be evaluated or extended across different platforms while sharing the same human-interactive scenarios and evaluation protocol.}
  \label{fig:platform-assets}
\end{figure}
\vspace{-1.0em}

Follow-Bench~2.0 provides reusable robot and sensor assets to support different embodied RPF policies under the same scenario and evaluation protocol. As shown in Figure~\ref{fig:platform-assets}, the benchmark includes multiple robot embodiments: a wheeled robot, a quadruped robot, and a humanoid robot. These assets allow researchers to evaluate whether a following policy is tied to a specific embodiment or can be transferred across platforms with different motion constraints, viewpoints, and sensor mounting positions. The wheeled robot serves as the default platform for controlled benchmark evaluation, while the quadruped and humanoid robots provide additional embodiments for studying more diverse following behaviours. Together with RGB-D cameras, LiDAR, and simulator-side visibility annotation, these assets make Follow-Bench~2.0 more than a fixed scenario collection: the same human-interactive environments can be reused for policy debugging, expert demonstration collection, imitation learning, and future extensions to new robot platforms.

\paragraph{Extension to guidance tasks.}
The modular design also supports guidance scenarios with limited changes. After importing a scene and extracting its occupancy map, a user can specify a global route for the robot and initialize the accompanied person behind it, thereby reversing the target--robot role used in person following. The existing pedestrian controllers, sensor interfaces, logging tools, and evaluation pipeline can then be reused; packaging this configuration as a standard task is planned for a future simulator release.

\FloatBarrier
\subsection{Implementation Details}
\label{app:impl-details}

\paragraph{Hardware.} All experiments are conducted on a single laptop equipped with an Intel Core Ultra~9 275HX CPU ($24$ cores), $30$\,GB RAM, and an NVIDIA GeForce RTX~5090 Laptop GPU ($24$\,GB VRAM). Unreal Engine rendering, the modular perception pipeline (MOT, Target-ReID, LiDAR clustering), all modular planners, and the inference of TrackVLA (Qwen3-4B language backbone with DINO and SiGLIP vision towers) and OA-VAT share this single GPU. 

\paragraph{Simulator and policy clock.} All runs use Unreal Engine in fixed-step synchronous mode at $\Delta t=0.1$\,s (10\,Hz), and every policy is invoked once per simulator step. The simulator waits for the current policy action before advancing, which keeps metric sampling frequency fixed across methods; wall-clock observation-to-action latency is reported separately.

\begin{table}[!htbp]
  \centering
  \caption{Per-scenario benchmark results on Follow-Bench~2.0 under back-following at $1.5$\,m, averaged over the four difficulty presets ($50$ trials per cell). Best per metric within each scenario block is bold-faced.}
  \label{tab:full-results-all-difficulties}
  \scriptsize
  \setlength{\tabcolsep}{4pt}
  \renewcommand{\arraystretch}{1.0}
  \resizebox{0.78\linewidth}{!}{%
  \begin{tabular}{l cc cc c}
    \toprule
    \textbf{Policy}
    & \textbf{SR$\uparrow$}  & \textbf{ASR$\uparrow$}
    & \textbf{TinTPerson$\uparrow$}
    & \textbf{TinPrivate$\downarrow$} & \textbf{Jerk$\downarrow$} \\
    & (\%) & (\%) & (\%) & (\%) & (m/s$^3$) \\
    \midrule
    \multicolumn{6}{l}{\emph{Open Plaza}} \\
    MPC w/ Traj. + KPR    & 23.5          & 23.5          & 91.4          & 36.6          & 41.8          \\
    MPC w/ Traj. + R18    & 25.0          & 30.5          & \textbf{97.8} & 38.1          & 29.8          \\
    MPC w/ DS. + R18      & 35.0          & \textbf{59.5} & 37.5          & \textbf{11.7} & 120.6         \\
    MPC + R18             & 16.1          & 16.6          & 72.7          & 41.1          & \textbf{24.1} \\
    TrackVLA              & 16.0          & 46.5          & 18.7          & 25.7          & 65.3          \\
    OA-VAT                & \textbf{40.0} & 58.5          & 10.9          & 18.0          & 61.6          \\
    \cmidrule(lr){1-6}
    \multicolumn{6}{l}{\emph{Street Corner}} \\
    MPC w/ Traj. + KPR    & \textbf{96.5} & 96.5          & 97.7          & 1.0           & 21.9          \\
    MPC w/ Traj. + R18    & 95.4          & 95.4          & \textbf{99.1} & 0.8           & 9.6           \\
    MPC w/ DS. + R18      & 90.0          & 94.5          & 34.7          & \textbf{0.3}  & 51.0          \\
    MPC + R18             & 93.9          & 94.9          & 43.2          & 1.5           & \textbf{4.8}  \\
    TrackVLA              & 32.0          & 85.5          & 12.3          & 1.0           & 32.2          \\
    OA-VAT                & 13.0          & \textbf{97.5} & 5.3           & 0.4           & 12.1          \\
    \cmidrule(lr){1-6}
    \multicolumn{6}{l}{\emph{Cluttered Park}} \\
    MPC w/ Traj. + KPR    & 0.0           & 49.0          & \textbf{80.9} & 3.6           & 82.7          \\
    MPC w/ Traj. + R18    & 0.0           & 55.5          & 79.6          & \textbf{3.2}  & 95.2          \\
    MPC w/ DS. + R18      & 0.0           & 59.5          & 39.9          & 4.5           & 103.3         \\
    MPC + R18             & 0.0           & 13.5          & 50.5          & 5.9           & \textbf{34.5} \\
    TrackVLA              & 0.0           & 55.3          & 22.1          & 9.2           & 51.6          \\
    OA-VAT                & 0.0           & \textbf{71.5} & 13.1          & 8.9           & 24.4          \\
    \cmidrule(lr){1-6}
    \multicolumn{6}{l}{\emph{Narrow Passage}} \\
    MPC w/ Traj. + KPR    & 83.0          & 88.0          & 88.1          & 1.4           & 15.9          \\
    MPC w/ Traj. + R18    & \textbf{86.9} & \textbf{91.5} & \textbf{94.7} & 1.6           & 13.3          \\
    MPC w/ DS. + R18      & 30.0          & 73.0          & 30.0          & \textbf{0.6}  & 66.6          \\
    MPC + R18             & 71.0          & 84.5          & 40.9          & 1.7           & \textbf{9.5}  \\
    TrackVLA              & 10.5          & 70.5          & 10.5          & \textbf{0.6}  & 32.7          \\
    OA-VAT                & 4.0           & 85.5          & 8.4           & 1.8           & 8.9           \\
    \midrule
    \multicolumn{6}{l}{\textbf{Average} (over the four scenarios and the four difficulty presets)} \\
    MPC w/ Traj. + KPR    & 50.8          & 64.2          & 89.5          & 10.6          & 40.6          \\
    MPC w/ Traj. + R18    & \textbf{51.6} & 68.1          & \textbf{92.8} & 11.0          & 37.1          \\
    MPC w/ DS. + R18      & 38.8          & 71.6          & 35.5          & \textbf{4.3}  & 85.4          \\
    MPC + R18             & 45.2          & 52.3          & 51.8          & 12.5          & \textbf{18.3} \\
    TrackVLA              & 14.6          & 64.5          & 15.9          & 9.2           & 45.4          \\
    OA-VAT                & 14.2          & \textbf{78.2} & 9.4           & 7.2           & 26.8          \\
    \bottomrule
  \end{tabular}%
  }
\end{table}

\FloatBarrier
\subsection{Author Teleoperation Reference}
\label{app:author-teleop}

Table~\ref{tab:author-teleop} reports an author-teleoperation reference for checking that the diagnostic scenarios remain solvable and that the difficulty presets are ordered as intended. Two co-author volunteers with robotics backgrounds each completed five trials in every combination of scenario, difficulty preset, and following configuration, for $160$ episodes per operator and $320$ episodes in total. The trials use the same pre-generated evaluation seeds as the automated policies, so actor appearance, start--goal configuration, and weather are those determined by the seed. Control is by keyboard on a Unitree G1. Each operator sees the onboard RGB image and a third-person view, and is asked to maintain the specified back- or side-following formation. Success uses the same collision and long-term target-loss criterion as the policy evaluation. The third-person view and the different embodiment mean that these rates are a solvability and difficulty check, not a perception-matched ceiling for the Scout-Mini policy results. Pooled success decreases from $78.8\%$ at \textit{easy} to $51.3\%$ at \textit{legend}. Back-following ($76.3\%$) remains higher than side-following ($55.6\%$), and no cell falls below $30\%$.

\begin{center}
  \begin{minipage}{\linewidth}
  \centering
  \captionsetup{hypcap=false}
  \captionof{table}{Pooled author-teleoperation success rate (\%) on Follow-Bench~2.0. Each cell pools ten trials, five from each of two co-author volunteers with robotics backgrounds, drawn from the shared evaluation seeds.}
  \label{tab:author-teleop}
  \scriptsize
  \setlength{\tabcolsep}{3pt}
  \renewcommand{\arraystretch}{1.0}
  \resizebox{0.6\linewidth}{!}{%
  \begin{tabular}{@{}ll cccc c@{}}
    \toprule
    \textbf{Scenario} & \textbf{Follow}
      & \textit{easy} & \textit{hard} & \textit{boss} & \textit{legend}
      & \textbf{Avg.} \\
    \midrule
    Open Plaza & back & 100.0 & 80.0 & 80.0 & 50.0 & 77.5 \\
               & side & 100.0 & 80.0 & 70.0 & 60.0 & 77.5 \\
    Street Corner & back & 90.0 & 90.0 & 90.0 & 70.0 & 85.0 \\
                  & side & 50.0 & 40.0 & 40.0 & 30.0 & 40.0 \\
    Cluttered Park & back & 90.0 & 80.0 & 70.0 & 70.0 & 77.5 \\
                   & side & 70.0 & 50.0 & 60.0 & 40.0 & 55.0 \\
    Narrow Passage & back & 70.0 & 70.0 & 80.0 & 40.0 & 65.0 \\
                   & side & 60.0 & 60.0 & 30.0 & 50.0 & 50.0 \\
    \midrule
    \textbf{All} & back & 87.5 & 80.0 & 80.0 & 57.5 & 76.3 \\
                 & side & 70.0 & 57.5 & 50.0 & 45.0 & 55.6 \\
                 & both & 78.8 & 68.8 & 65.0 & 51.3 & 65.9 \\
    \bottomrule
  \end{tabular}%
  }
  \end{minipage}
\end{center}

\end{document}